\documentclass{article} 
\usepackage{iclr2027_conference,times}
\usepackage[table]{xcolor}
\usepackage{textcomp}

\usepackage{amsmath,amsfonts,bm}

\def\eqref#1{equation~\ref{#1}}

\def\1{\bm{1}}

\DeclareMathAlphabet{\mathsfit}{\encodingdefault}{\sfdefault}{m}{sl}
\SetMathAlphabet{\mathsfit}{bold}{\encodingdefault}{\sfdefault}{bx}{n}

\usepackage{hyperref}
\usepackage{url}
\usepackage{makecell}
\usepackage{threeparttable}
\usepackage{wrapfig}
\usepackage{marvosym}
\usepackage{multirow}
\usepackage{booktabs}
\usepackage{subfigure}
\usepackage{graphicx}

\title{
SAGE: Salient Factor Discovery and Generation with Visual Foundation Representations}

\iclrfinalcopy
\AddToHook{cmd/maketitle/after}{\lhead{Preprint}}

\author{%
\textbf{Shuang Liang}$^{1,2\dagger}$\quad
\textbf{Lejun Liao}$^{2\dagger}$\quad
\textbf{Shiyuan Zhang}$^{2,3\dagger}$\quad
\textbf{Max C. Zhang}$^{2}$\\
\textbf{Xiaolong Luo}$^{4}$\quad 
\textbf{Han Wang}$^{1}$ \quad
\textbf{Stefano Anzellotti}$^{2}$\quad
\textbf{Yuan Yuan}$^{2}$\textsuperscript{\Letter}\\[3pt]
$^{1}$HKU\quad
$^{2}$Boston College\quad
$^{3}$University of Virginia\quad 
$^{4}$Harvard University\quad
}

\begin{document}

\newcommand{\sz}[1]{\textcolor{red}{[SZ: #1]}}
\newcommand{\tsl}[1]{\textcolor{blue}{[SL: #1]}}

\maketitle
\begingroup
\renewcommand{\thefootnote}{}
\footnotetext{
$^\dagger$These authors contributed equally. 
\textsuperscript{\Letter}Corresponding author: Yuan Yuan (\href{mailto:yuanyua@bc.edu}{\nolinkurl{yuanyua@bc.edu}}).}
\endgroup

\begin{abstract}

Given a target dataset, such as faces with eyeglasses, and a background dataset, such as faces without, contrastive analysis separates \textit{salient} factors specific to the target from \textit{common} content shared by both. We aim for salient representations that capture target-specific detail in each image, such as the shape, color, and position of the glasses, so that they reveal subtypes without subtype labels and guide the generation of new examples of a discovered subtype, even one with no name or text description. We introduce SAGE, which learns both factors directly in the high-dimensional spatial latent of a frozen representation autoencoder and conditions a diffusion transformer on the learned salient representation of a reference image. On Digits-ImageNet and FFHQ eyeglasses, SAGE combines high-fidelity \textit{reconstruction} (rFID below $2$) with unsupervised \textit{subtype discovery}, recovering the digits better than baselines (probe accuracy $0.950$ vs.\ at most $0.281$) and revealing eyewear types, finer sunglasses styles, and mislabeled images; salient-conditioned \textit{generation} raises Digits-ImageNet subtype accuracy over the unfactorized latent ($90.5\%$ vs.\ $27.7\%$) and diversity on both datasets. On retinal OCT, SAGE's salient space separates three diseases using only normal/disease labels.

\end{abstract}

\begin{figure*}[h]
    \centering
    \includegraphics[width=\linewidth]{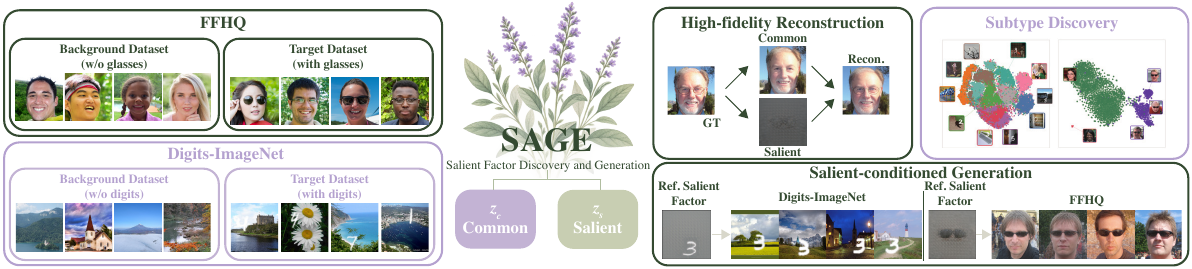}
    \caption{\textbf{Overview of SAGE.}
    Given background (BG) and target (TG) datasets, SAGE uses only dataset labels to decompose frozen self-supervised visual representations into \textit{common} and \textit{salient} factors, capturing shared content and target-specific variation, respectively. The decomposition supports high-fidelity reconstruction, subtype discovery in the salient space, and salient-conditioned generation that preserves target attributes while varying common content.}
    \label{fig:overview}
\end{figure*}

\section{Introduction}

Learning what two data distributions share and what sets one apart is central to multimodal learning~\citep{dufumier2025align}, domain adaptation~\citep{lee2021dranet}, and disentanglement~\citep{sanchez2020learning}. Contrastive analysis (CA) poses this problem with dataset labels alone: given a target and a background dataset, it separates \emph{salient} factors $z_s$ specific to the target from \emph{common} factors $z_c$ shared by both. For faces with and without eyeglasses as target and background (Fig.~\ref{fig:overview}), $z_s$ should capture eyewear-related variation (e.g., frame style and color) and $z_c$ the content both groups share (e.g., identity, pose, and scene). The stakes are especially high in medicine, where clinical group labels may be available even when the corresponding imaging differences remain unknown. Autism, for instance, is diagnosed from behavior, which tells us who is affected but not how their brains differ; contrasting brain MRI of the autism and control groups can reveal these differences~\citep{aglinskas2022contrastive}. Such patterns are subtle, hard to describe in words, and not known in advance, yet finding them could provide objective imaging markers, deepen the understanding of a disease, and even uncover its subtypes. All of these settings raise the same question:

\vspace{-3pt}{\centering\colorbox{orange!10}{\parbox{0.86\linewidth}{\centering\itshape Can we learn salient representations that capture target-specific detail in each image, reveal subtypes without labels, and generate new examples of what they discover?}}\par}\vspace{-1pt}

Answering it requires a representation that is semantically rich enough for discovery and decodable back to images, and existing CA methods offer complementary strengths toward these goals: generative CA models~\citep{abid2019contrastive,louiset2023sepvae,carton2024double} decode their factors but reconstruct complex natural images poorly, and contrastive CA methods~\citep{louiset2024separating} learn semantically rich factors but have no decoder. In this work, we propose \textbf{SAGE} (\textbf{Sa}lient factor discovery and \textbf{Ge}neration), a contrastive analysis approach that factorizes the frozen spatial latent $z$ of a representation autoencoder (RAE), RAEv2~\citep{singh2026improved}, into $z_c$ and $z_s$, whose sum the frozen RAE decoder maps back to images. Built on a pretrained foundation encoder, DINOv3~\citep{simeoni2025dinov3}, this latent is semantically rich, keeps a high-dimensional spatial layout in which salient factors can retain instance-level detail such as the shape, color, and position of a particular pair of glasses, decodes with high fidelity, and supports state-of-the-art diffusion generation~\citep{zheng2025diffusion,singh2026improved}.

This latent, however, mixes target-specific variation with all other image content~\citep{li2023addressing}. Separating the two is nontrivial: reconstruction alone is already satisfied by an empty $z_s$, a high-dimensional $z_s$ can easily absorb shared content, and it can capture only a generic template of the attribute while instance-specific detail stays in $z_c$. SAGE counters these failures with objectives that include swapping salient factors between background and target images, keeping them sparse, and requiring them to be recoverable after the swap. The learned salient factors cleanly separate target-specific content and reveal subtypes, and they can be swapped or removed to edit individual images. SAGE can further generate images. Because the salient variation is often not known in advance and has no name or text description, instruction- or prompt-conditioned generation~\citep{liu2023visual,wu2025qwen} cannot even be applied; instead, in a second stage, SAGE conditions a diffusion transformer on the learned salient representation of a reference image to generate new images that keep its salient subtype while their common content varies. This in turn indicates that the salient condition carries little of the reference's shared content.

Our contributions answer each part of the question above: (1)~\textbf{Learning clean, high-fidelity salient factors}: salient factors decode to only the digit or eyeglasses, and together with the common factors they reconstruct Digits-ImageNet and FFHQ images with rFID below $2$, versus above $120$ for generative CA baselines. (2)~\textbf{Revealing subtypes without labels}: on Digits-ImageNet, the salient space recovers the ten digits far better than a contrastive CA baseline on the same encoder (probe accuracy $0.950$ vs.\ $0.148$); on FFHQ, it separates eyewear types, finer sunglasses styles, and a mislabeled cluster; and on retinal OCT, it distinguishes three diseases. (3)~\textbf{Generating what is discovered}: on Digits-ImageNet, conditioning on a reference's salient representation, without text prompts or subtype labels, preserves its subtype far more often and produces more varied samples than the unfactorized latent (subtype accuracy $90.5\%$ vs.\ $27.7\%$; Vendi $12.50$ vs.\ $2.65$).

\section{Related Work}
\label{sec:related}

\paragraph{Contrastive analysis.}
Contrastive analysis (CA) separates factors shared by background and target datasets from target-specific ones using only dataset labels, with applications in neuroimaging~\citep{aglinskas2022contrastive,zhu2026deepcor} and disease subgroup discovery~\citep{louiset2026automatic}. cVAE~\citep{abid2019contrastive}, SepVAE~\citep{louiset2023sepvae}, and Double-InfoGAN~\citep{carton2024double} learn the factors with a generator trained from scratch; SepCLR~\citep{louiset2024separating} learns strong salient representations under the InfoMax principle but has no decoder, and adding a reconstruction objective weakens its factor separation. CS-StyleGAN~\citep{he2025learning} separates factors in an inverted StyleGAN latent and refines reconstructions with input features, and the concurrent Diff-CA~\citep{soumm2026diffca} decomposes compact conditioning tokens of a diffusion generator. These methods represent the salient factor as a low-dimensional vector or global token; SAGE factorizes the full spatial latent of a frozen RAE and uses it for both discovery and generation.

\paragraph{Representation autoencoders and conditioned generation.}
Recent work replaces the VAE of latent diffusion with frozen pretrained encoders~\citep{zheng2025diffusion,tong2026scaling,shi2025latent,gao2026one}; RAEs pair such encoders with ViT decoders that match or exceed standard VAEs~\citep{zheng2025diffusion}, and RAEv2 adds REPA~\citep{singh2026improved,yu2024representation}. SAGE factorizes this latent rather than only generating in it. Representation-conditioned generation~\citep{li2024return} conditions on a full self-supervised representation, and diffusion editing~\citep{meng2022sdedit,wu2025qwen} changes prompt-specified attributes; SAGE instead conditions on a discovered salient representation, specifying a subtype by example.

\section{Methodology}
\label{sec:methodology}

\begin{figure*}[t]
    \centering
    \includegraphics[width=0.7\linewidth]{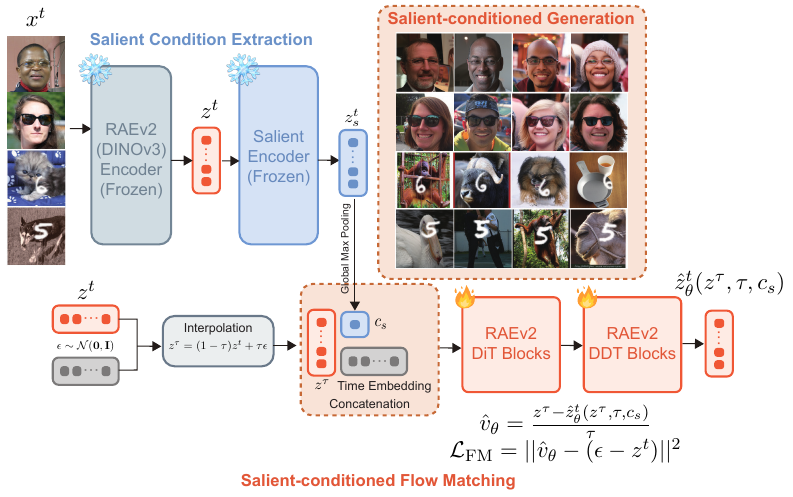}
    \caption{\textbf{Stage~2 of SAGE for salient-conditioned generation.}
    The frozen RAEv2 and salient encoders extract a reference salient representation, max-pooled to $c_s$. An RAEv2-initialized diffusion transformer is trained with conditional flow matching in this latent. Top right: generated samples; each row keeps the eyewear type or digit of the reference at left while faces and scenes vary.}
    \label{fig:stage-2-framework}
\end{figure*}

Given background and target datasets $\mathcal{D}_b$ and $\mathcal{D}_t$, SAGE uses only dataset labels $d\in\{b,t\}$, not subtype labels, to factorize the latent $z=E(x)$ of a frozen RAE with decoder $D$. Stage~1 learns the factorization; Stage~2 freezes it and trains a salient-conditioned diffusion transformer (Figs.~\ref{fig:stage-1-framework},~\ref{fig:stage-2-framework}).

\subsection{Salient Factor Discovery}
\label{sec:stage1_method}
\begin{figure*}[t]
    \centering
    \includegraphics[width=0.74\linewidth]{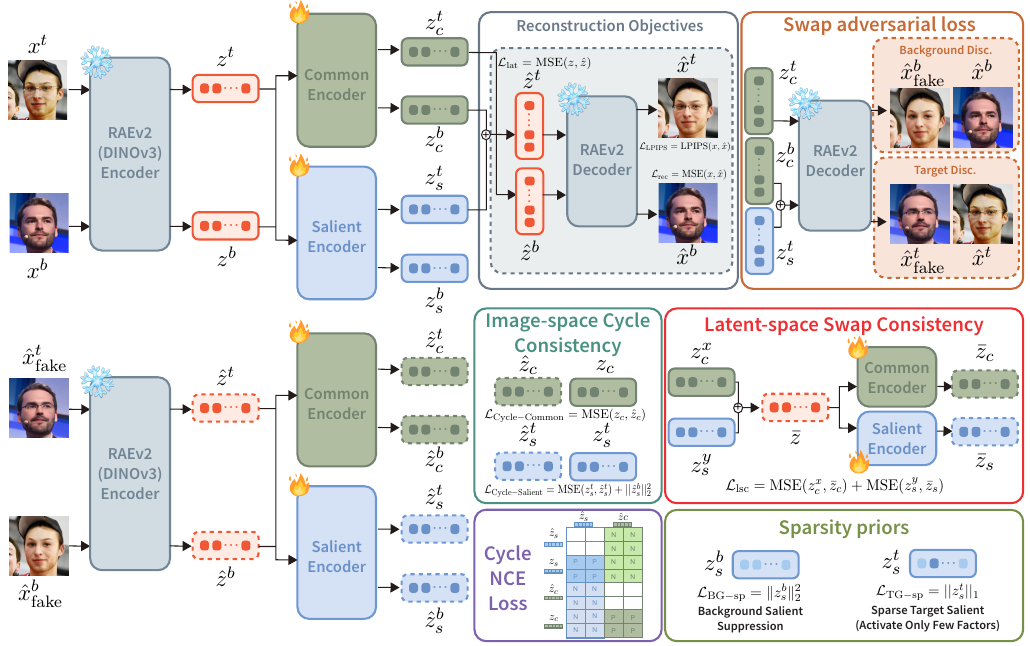}
    \caption{\textbf{Stage~1 of SAGE for salient factor discovery.}
    Trainable common and salient encoders split latents of the frozen RAEv2 encoder (DINOv3); the frozen RAEv2 decoder maps factors back to images. Top: reconstruction ($\mathcal{L}_{\mathrm{rec}}$) and swap adversarial training ($\mathcal{L}^{G}_{\mathrm{swap}}$). Bottom: swapped images are re-encoded for $\mathcal{L}_{\mathrm{cyc}}$ and $\mathcal{L}_{\mathrm{cNCE}}$, while $\mathcal{L}_{\mathrm{lsc}}$ and the sparsity priors act directly on the factors. Snowflakes and flames mark frozen and trainable modules.}
    \label{fig:stage-1-framework}
\end{figure*}

\paragraph{Why a frozen RAE latent space.}
SAGE needs a latent that is semantically expressive for discovery and decodable for reconstruction, manipulation, and generation. Unlike the VAE latents of standard latent diffusion, which are trained for reconstruction and carry limited semantic structure~\citep{zheng2025diffusion}, RAEs pair frozen pretrained visual encoders with trained decoders, providing both properties and supporting high-quality latent diffusion~\citep{zheng2025diffusion,singh2026improved}. We use the frozen RAEv2 encoder and decoder~\citep{singh2026improved}: $E$ is DINOv3-L~\citep{simeoni2025dinov3} with multi-layer summation (MLS), and $D$ reconstructs images from $z$. We factorize this spatial latent directly, without a learned bottleneck, to retain localized salient content (Fig.~\ref{fig:factor-decoding-benchmark}); freezing $E$ and $D$ attributes decoded changes to the learned factors and lets Stage~2 reuse RAEv2's pretrained diffusion transformer.

\paragraph{Common and salient encoders.}
For $z=E(x)\in\mathbb{R}^{N\times C}$, trainable encoders produce same-shape common and salient factors $z_c=E_c(z)$ and $z_s=E_s(z)$, decoded as $D(z_c,z_s):=D(z_c+z_s)$; the sum stays in the frozen decoder's latent space, and $z_s=\mathbf{0}$ means no target-specific content. We require the factors to be decodable and composable, not statistically independent (Appendix~\ref{app:limitations}). Because dataset labels alone underdetermine this split, the objectives below reconstruct each image, push target-specific content into $z_s$, and preserve its image-specific appearance (Fig.~\ref{fig:stage-1-framework}).

\paragraph{Asymmetric reconstruction.}
Because the salient factor should carry only target-specific content, it is absent for background images but complements the common factor for targets: $\hat{x}^b=D(z_c^b,\mathbf{0})$ and $\hat{x}^t=D(z_c^t,z_s^t)$.
With $\ell(x,\hat{x})=\lVert x-\hat{x}\rVert_2^2+\mathrm{LPIPS}(x,\hat{x})$, we minimize
\begin{equation}
\mathcal{L}_{\mathrm{rec}}
=\mathbb{E}_{x^b}\!\left[\ell(x^b,\hat{x}^b)+\left\lVert z_c^b-z^b\right\rVert_2^2\right]
+\mathbb{E}_{x^t}\!\left[\ell(x^t,\hat{x}^t)+\left\lVert z_c^t+z_s^t-z^t\right\rVert_2^2\right].
\end{equation}
The latent terms tie decoder inputs to real latents, avoiding plausible but unreliable compositions.

\paragraph{Swap adversarial loss.}
Reconstruction leaves factor allocation underdetermined: for a target image, the trivial split $z_c^t=z^t,\ z_s^t=\mathbf{0}$ reconstructs perfectly. To move target-specific content to $z_s$, we zero-initialize $E_s$ so that $z_s$ starts empty and acquires content only when required, then swap target salient factors. Swapping exploits dataset labels: if $z_s^t$ captures target-specific content, erasing it should yield a background-like image, while adding it to $z_c^b$ should yield a target-like image: $\hat{x}^{b}_{\mathrm{fake}}=D(z_c^t,\mathbf{0})$ and $\hat{x}^{t}_{\mathrm{fake}}=D(z_c^b,z_s^t)$. Because swapped images have no paired ground truth, background and target discriminators $\Delta_b$ and $\Delta_t$ distinguish $\hat{x}^{b}_{\mathrm{fake}}$ and $\hat{x}^{t}_{\mathrm{fake}}$ from reconstructions $\hat{x}^b$ and $\hat{x}^t$, respectively~\citep{goodfellow2014generative} (architecture in Appendix~\ref{app:stage1_architectures}); they minimize the hinge loss $\mathcal{L}^{D}_{\mathrm{swap}}$ (Appendix~\ref{app:swap_objective}). Using reconstructions as real samples prevents the discriminators from exploiting decoder artifacts to distinguish swapped outputs from input images. The encoders minimize the following loss, where $\sigma$ is the sigmoid function:
\begin{equation}
\label{eq:swap-generator-loss}
\mathcal{L}^{G}_{\mathrm{swap}}
=-\mathbb{E}\!\left[\log\sigma\!\left(\Delta_b(\hat{x}^{b}_{\mathrm{fake}})\right)\right]
-\mathbb{E}\!\left[\log\sigma\!\left(\Delta_t(\hat{x}^{t}_{\mathrm{fake}})\right)\right],
\end{equation}

\paragraph{Sparsity priors.}
Swap alone does not exclude common content from $z_s$. Since reconstruction discards $z_s^b$, we shrink background salient factors to zero with an $\ell_2^2$ penalty; for targets, an $\ell_1$ penalty favors sparse $z_s^t$ to keep common content, such as the face wearing glasses, out of it:
\begin{equation}
\mathcal{L}_{\mathrm{BG\text{-}sp}}
=\mathbb{E}_{x^b}\!\left[\left\lVert z_s^b\right\rVert_2^2\right],
\qquad
\mathcal{L}_{\mathrm{TG\text{-}sp}}
=\lambda\,\mathbb{E}_{x^t}\!\left[\left\lVert z_s^t\right\rVert_1\right].
\end{equation}
The dataset-specific $\lambda$ balances sparsity against retaining target detail (Appendix~\ref{app:stage1_training}).

\paragraph{Consistency objectives.}
Dataset-level adversarial matching cannot ensure $z_s^t$ carries instance-specific target content: a generic eyeglass prototype could pass both discriminators while instance-specific details remain in $z_c^t$. We therefore require the factors to be recoverable after swapping, encouraging each salient representation to be instance-specific; the image-space cycle loss re-encodes swapped images, $\hat{z}^t=E(\hat{x}^t_{\mathrm{fake}})$ and $\hat{z}^b=E(\hat{x}^b_{\mathrm{fake}})$, to recover their factors~\citep{zhu2017unpaired}:
\begin{equation}
\mathcal{L}_{\mathrm{cyc}}
=\mathbb{E}\Big[
\lVert E_c(\hat{z}^t)-z_c^b\rVert_2^2
+\lVert E_s(\hat{z}^t)-z_s^t\rVert_2^2
+\lVert E_c(\hat{z}^b)-z_c^t\rVert_2^2
+\lVert E_s(\hat{z}^b)\rVert_2^2
\Big],
\end{equation}
With the original factors as fixed targets (Appendix~\ref{app:consistency_objectives}), this loss checks rendered transfers and erasures; its final term enforces no salient content after erasure. The cheaper $\mathcal{L}_{\mathrm{lsc}}$ applies the same recovery in latent space, without decoding: for a random permutation $\pi$ of the minibatch,
\begin{equation}
\mathcal{L}_{\mathrm{lsc}}
=\mathbb{E}_i\Big[
\lVert E_c(z_c^i+z_s^{\pi(i)})-z_c^i\rVert_2^2
+\lVert E_s(z_c^i+z_s^{\pi(i)})-z_s^{\pi(i)}\rVert_2^2
\Big].
\end{equation}

\paragraph{Cycle NCE loss.}
Recovery losses do not preclude overlapping common and salient content, such as facial appearance present in both. Cycle NCE applies InfoNCE~\citep{oord2018representation} to globally pooled, $\ell_2$-normalized representations before and after swap--re-encoding, using the cycle to form positive pairs without augmentations. Because $\hat{x}^t_{\mathrm{fake}}=D(z_c^b,z_s^t)$ combines factors from different images, its re-encoded factors should recover $z_c^b$ and $z_s^t$ despite the changed context. For each factor, minibatch representations of the other factor, original or re-encoded, serve as negatives; same-factor representations are excluded, so the loss separates the two factors without pushing apart images within one, preserving subtype grouping (Appendix~\ref{app:cycle_nce}).

\paragraph{Full objective.}
The encoders minimize the unit-weighted sum,
\begin{equation}
\mathcal{L}_{\mathrm{SAGE}}=
\mathcal{L}_{\mathrm{rec}}
+\mathcal{L}^{G}_{\mathrm{swap}}
+\mathcal{L}_{\mathrm{BG\text{-}sp}}
+\mathcal{L}_{\mathrm{TG\text{-}sp}}
+\mathcal{L}_{\mathrm{cyc}}
+\mathcal{L}_{\mathrm{lsc}}+\mathcal{L}_{\mathrm{cNCE}},
\end{equation}
where $\lambda$ enters only through $\mathcal{L}_{\mathrm{TG\text{-}sp}}$. The discriminators minimize $\mathcal{L}^{D}_{\mathrm{swap}}$, and the two updates alternate at each step; the RAEv2 encoder and decoder remain frozen throughout.

\subsection{Salient-Conditioned Generation}
\label{sec:stage2_method}
After Stage~1, the frozen RAEv2 and salient encoders map a target image $x^t$ to $z_s^t=E_s(E(x^t))$; the common encoder is unnecessary. This representation conditions Stage~2 generation (Fig.~\ref{fig:stage-2-framework}): samples retain the reference's salient characteristics while their common content varies. We condition on $z_s$ rather than the full latent $z$, which also encodes common content such as the face or scene that generation would otherwise reproduce. Channel-wise max pooling gives $c_s=\mathrm{GMP}(z_s^t)\in\mathbb{R}^{C}$, discarding location but keeping channels active in few tokens of the sparse map; $c_s$ is projected to one token and concatenated with the noisy latent and time embeddings.

A diffusion transformer (DiT) with RAEv2's DDT head generates latents. Initialized from the pretrained RAEv2 checkpoint, it is trained with flow matching~\citep{lipman2023flow} in the same latent space on target images only. With $\epsilon\sim\mathcal{N}(\mathbf{0},\mathbf{I})$ and $\tau\in[0,1]$, let $z^\tau=(1-\tau)z^t+\tau\epsilon$. The DiT $F_\theta$ predicts the clean latent, giving $\hat{v}_\theta=\big(z^\tau-F_\theta(z^\tau,\tau,c_s)\big)/\max(\tau,\tau_{\min})$, and is trained to match the flow velocity by minimizing $\mathcal{L}_{\mathrm{fm}}=\mathbb{E}_{x^t,\epsilon,\tau}\lVert\hat{v}_\theta(z^\tau,\tau,c_s)-(\epsilon-z^t)\rVert_2^2$.
During training, a learnable null token replaces $c_s$ with probability $p_{\mathrm{uncond}}$, enabling classifier-free guidance~\citep{ho2022classifier} and unconditional sampling. At inference, we integrate from Gaussian noise at $\tau=1$ to $\tau=0$ conditioned on $c_s$ and decode with frozen $D$: different noise samples vary common content, while $c_s$ fixes salient characteristics (Appendix~\ref{app:stage2_details}).

\section{Experiments}
\label{sec:experiments}

\subsection{Experimental Settings}
\paragraph{Datasets \& metrics.}
We evaluate on \textbf{\textit{Digits-ImageNet}}, ImageNet images with and without an overlaid MNIST digit (ten digit subtypes)~\citep{lecun1998gradient,deng2009imagenet}; \textbf{\textit{FFHQ}} faces with and without eyeglasses (reading glasses or sunglasses)~\citep{karras2019style,ffhqfeatures}; and \textbf{\textit{OCT-Kermany}} normal retinal scans versus scans of three diseases (CNV, DME, and DRUSEN)~\citep{kermany2018identifying}. Subtype labels are used only for evaluation (Appendix~\ref{app:dataset_definitions}). We measure \textit{reconstruction} by rFID~\citep{heusel2017gans}, PSNR, SSIM~\citep{wang2004image}, and LPIPS~\citep{zhang2018unreasonable}; \textit{subtype discovery} by Salient LP and ARI~\citep{hubert1985comparing}/NMI~\citep{strehl2002cluster} of $k$-means clusters; and \textit{generation} by gFID~\citep{heusel2017gans}, generated-image subtype accuracy, and within-condition Vendi score~\citep{friedman2023vendi} (Appendices~\ref{app:linear_probe}, \ref{app:ari_protocol}, and~\ref{app:within_condition_diversity}).

\paragraph{Baselines and implementation.}
We compare with cVAE, SepVAE, Double-InfoGAN, and SepCLR~\citep{abid2019contrastive,louiset2023sepvae,carton2024double,louiset2024separating}, as well as DINOv3 + SepCLR and unfactorized DINOv3 features. Full implementation details appear in Appendices~\ref{app:baselines}, \ref{app:stage1_details}, and~\ref{app:stage2_details}.

\subsection{Salient Factor Discovery}
\label{sec:salient_experiments}

\begin{table}[!t]
\centering
\setlength{\abovecaptionskip}{0pt}
\caption{\textbf{Benchmark comparison on Digits-ImageNet and FFHQ.} rFID: reconstruction Fr\'echet Inception Distance; N/A: no decoder; Double-InfoGAN uses its native $128\times128$ resolution; all other methods use $256\times256$. BG/TG Sil.: silhouette score of background versus target salient representations (Appendix Table~\ref{tab:silhouette_scores}). LP Acc.: linear-probe subtype accuracy, five-fold on Digits-ImageNet and balanced on FFHQ (Appendix~\ref{app:linear_probe}). ARI/NMI: $k$-means on the $\ell_2$-normalized, unprojected salient representations, with $k$ set to the subtype count (Appendix~\ref{app:ari_protocol}). FFHQ labels are audit-corrected.}
\label{tab:benchmark_main}
\setlength{\tabcolsep}{4.5pt}
\renewcommand{\arraystretch}{1.15}
\resizebox{\linewidth}{!}{%
\begin{tabular}{l cccc ccc cccc ccc}
\toprule
& \multicolumn{7}{c}{\textbf{Digits-ImageNet}}
& \multicolumn{7}{c}{\textbf{FFHQ}} \\
\cmidrule(lr){2-8}
\cmidrule(lr){9-15}
& \multicolumn{4}{c}{Reconstruction} & \multicolumn{3}{c}{Salient representation}
& \multicolumn{4}{c}{Reconstruction} & \multicolumn{3}{c}{Salient representation} \\
\cmidrule(lr){2-5}\cmidrule(lr){6-8}\cmidrule(lr){9-12}\cmidrule(lr){13-15}
Method
& rFID\,\textdownarrow & PSNR\,\textuparrow & SSIM\,\textuparrow & LPIPS\,\textdownarrow & BG/TG Sil.\,\textuparrow & LP Acc.\,\textuparrow & ARI/NMI\,\textuparrow
& rFID\,\textdownarrow & PSNR\,\textuparrow & SSIM\,\textuparrow & LPIPS\,\textdownarrow & BG/TG Sil.\,\textuparrow & LP Acc.\,\textuparrow & ARI/NMI\,\textuparrow \\
\midrule
cVAE~{\scriptsize\citep{abid2019contrastive}}
& 151.0 & 18.33 & 0.478 & 0.673 & 0.006 & 0.135 & 0.001/0.003
& 188.2 & 18.42 & 0.567 & 0.590 & 0.027 & 0.898 & 0.001/0.001 \\
SepVAE~{\scriptsize\citep{louiset2023sepvae}}
& 162.5 & 17.50 & 0.461 & 0.696 & 0.006 & 0.145 & 0.009/0.019
& 197.4 & 17.67 & 0.552 & 0.598 & 0.206 & 0.897 & -0.001/0.000 \\
Double-InfoGAN~{\scriptsize\citep{carton2024double}}
& 171.4 & 15.97 & 0.375 & 0.668 & 0.013 & 0.281 & 0.001/0.003
& 122.5 & 16.73 & 0.424 & 0.458 & 0.050 & 0.964 & 0.298/0.299 \\
SepCLR~{\scriptsize\citep{louiset2024separating}}
& N/A & N/A & N/A & N/A & 0.546 & 0.180 & 0.000/0.001
& N/A & N/A & N/A & N/A & 0.752 & 0.970 & 0.822/0.696 \\
DINOv3 + SepCLR
& N/A & N/A & N/A & N/A & 0.642 & 0.148 & 0.000/0.001
& N/A & N/A & N/A & N/A & 0.551 & 0.976 & 0.915/0.825 \\
\midrule
\rowcolor{gray!15}\textbf{SAGE (Ours)}
& \textbf{1.78} & \textbf{20.56} & \textbf{0.571} & \textbf{0.216} & \textbf{0.820} & \textbf{0.950} & \textbf{0.337/0.472}
& \textbf{1.65} & \textbf{23.02} & \textbf{0.723} & \textbf{0.163} & \textbf{0.873} & \textbf{0.983} & \textbf{0.939/0.865} \\
\bottomrule
\end{tabular}%
}
\end{table}

We first ask whether SAGE isolates a clean salient factor, one that contains the target content and little else, while the two factors together still reconstruct the image.

\begin{figure}[b]
\centering
\includegraphics[width=0.72\linewidth]{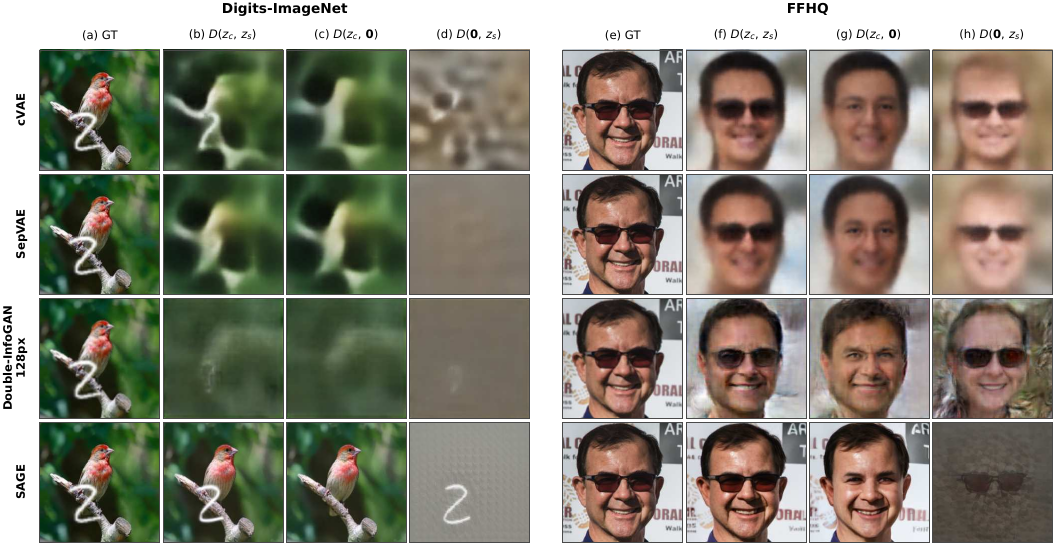}
\caption{\textbf{Reconstruction, common-only, and salient-only decoding.}
Columns show the input, the full reconstruction $D(z_c,z_s)$, common-only decoding $D(z_c,\mathbf{0})$, and salient-only decoding $D(\mathbf{0},z_s)$ on Digits-ImageNet (left) and FFHQ (right); rows compare cVAE, SepVAE, Double-InfoGAN, and SAGE. A clean salient factor renders only the digit or eyeglasses.}
\label{fig:factor-decoding-benchmark}
\end{figure}

\paragraph{Qualitative results.}
Among decoder-based methods, only SAGE separates the target attribute cleanly (Fig.~\ref{fig:factor-decoding-benchmark}; additional randomly selected examples in Appendix Fig.~\ref{fig:factor-decoding-supp}): it reconstructs scenes and faces with their digits or eyeglasses, common-only decoding removes only the target attribute, and salient-only decoding renders the digit or eyeglasses alone at their original location and shape. Generative baselines blur reconstructions and alter the scene or face even in common-only decoding; their salient-only outputs show no recognizable digit on Digits-ImageNet and blurred eyeglasses with much of the face on FFHQ, indicating common-content leakage. At the dataset level, SAGE's salient t-SNE separates background from target images (Fig.~\ref{fig:tsne-salient-atlas}a, d), as the sparsity prior drives background salient factors toward zero; those of cVAE and, on Digits-ImageNet, SepVAE mix the two (Appendix Fig.~\ref{fig:tsne_baselines}).

\paragraph{Quantitative results.}
SAGE achieves rFID $1.78$ on Digits-ImageNet and $1.65$ on FFHQ, whereas generative baselines trained from scratch exceed $120$ (Table~\ref{tab:benchmark_main}), because SAGE factorizes a frozen RAEv2 latent and decodes it with the pretrained decoder, adding little error over decoding $z$ directly ($1.78$ versus $0.39$ on Digits-ImageNet). Its salient space also separates background from target images best among all methods (silhouette $0.820$ and $0.873$). Linear probes (Appendix~\ref{app:full_ablation}) show that digit identity is far more accessible in $z_s$ than in $z$ ($0.950$ versus $0.328$), while ImageNet category, a property of the shared scene, is readily decoded from $z$ and $z_c$ ($0.765$) but hardly from $z_s$ ($0.045$). SepCLR and DINOv3 + SepCLR lack decoders; their comparison follows.

\subsection{Subtype Discovery}
\label{sec:stage1_experiments}

\begin{figure}[t]
\centering
\includegraphics[width=0.85\textwidth]{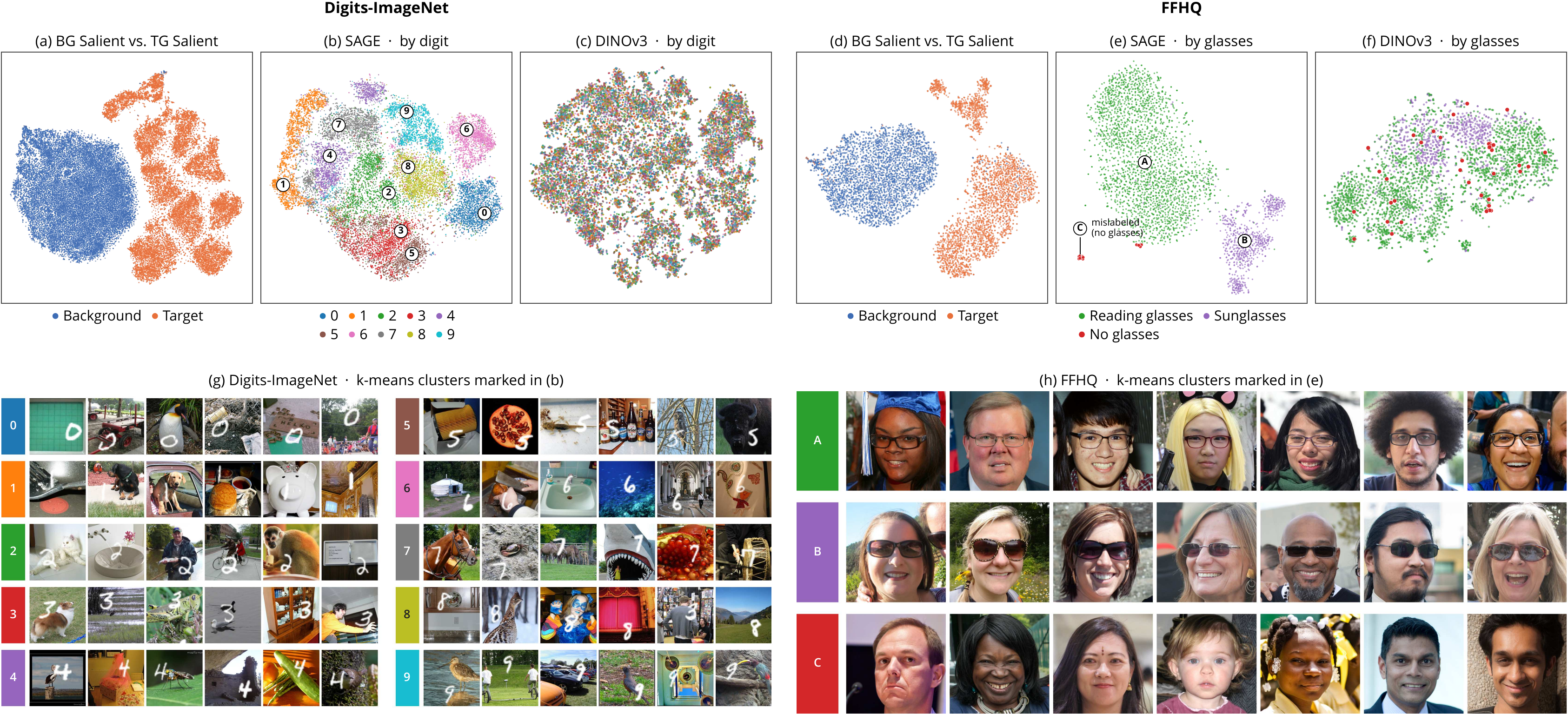}
\caption{\textbf{Subtype discovery in the salient space.} Top: t-SNE of SAGE's salient representations separates background and target images (a, d) and organizes digit and eyewear subtypes (b, e), whereas raw DINOv3 features mix them (c, f). Bottom: examples from the $k$-means clusters marked in (b, e) show coherent digit and eyewear groups (g, h); FFHQ cluster C contains images mislabeled as wearing glasses, and red points are audited no-glasses images.}
\label{fig:tsne-salient-atlas}
\end{figure}

We next ask whether the salient factor discovers target subtypes without subtype labels, both the expected ten digits and two eyewear types and groups we did not expect.

\paragraph{Qualitative results.}
Figure~\ref{fig:tsne-salient-atlas} shows t-SNE~\citep{van2008visualizing} of salient representations, colored by dataset label and, for targets, subtype labels used only for visualization. Background--target separation (Fig.~\ref{fig:tsne-salient-atlas}a, d; Section~\ref{sec:salient_experiments}) does not imply subtype separation: SepCLR and DINOv3 + SepCLR also separate the datasets, yet their target digits remain mixed (Appendix~\ref{app:tsne_baselines}). SAGE recovers the expected subtypes: the ten digit identities form locally coherent, partly overlapping groups, and reading glasses and sunglasses form two separate groups (Fig.~\ref{fig:tsne-salient-atlas}b, e); raw $z$ separates neither (Fig.~\ref{fig:tsne-salient-atlas}c, f). Samples from each group (Fig.~\ref{fig:tsne-salient-atlas}g, h) share the digit or eyewear type while their scenes and faces differ, so the grouping follows the target attribute rather than shared content.

\paragraph{Discovering finer eyewear styles.}
Within the sunglasses group (Fig.~\ref{fig:tsne-salient-atlas}e), SAGE separates three styles: thick, angular wayfarer-like frames; thin-rimmed aviator-like lenses; and a broader mixed group (Appendix Fig.~\ref{fig:sunglasses-subtypes}). The names describe the samples; the data have no style labels.

\paragraph{Discovering mislabeled images.}
Beyond reading glasses and sunglasses, the FFHQ t-SNE contains cluster C (Fig.~\ref{fig:tsne-salient-atlas}e, h), whose images lack eyeglasses but whose automatic annotations label them as wearing glasses, often because of face paint or hat-brim shadows (Appendix Fig.~\ref{fig:ffhq-noglasses-failures}c). This cluster led us to audit all 3,437 target validation images: annotators labeled the visible eyewear without seeing SAGE's assignments and corrected 125 labels (3.64\%; all corrections are shown in Appendix Fig.~\ref{fig:azure_label_errors}), including 39 no-glasses images; SAGE places 34 of these 39 in cluster C, while the other five show face paint or masks around the eyes (Appendix Fig.~\ref{fig:ffhq-noglasses-failures}a, b). With $k=3$, including no-glasses images, SAGE reaches ARI/NMI $0.937/0.865$, whereas every baseline remains below $0.46/0.56$; e.g., DINOv3 + SepCLR drops from $0.915/0.825$ to $0.453/0.550$ (Appendix Table~\ref{tab:ari_spaces}). SAGE thus also exposes label noise; Table~\ref{tab:benchmark_main} uses the audit-corrected labels for all methods.

\paragraph{Quantitative results.}
On Digits-ImageNet, SAGE reaches digit probe accuracy $0.950$, versus at most $0.281$ for baselines (Table~\ref{tab:benchmark_main}) and $0.328$ for the unfactorized RAEv2 latent $z$ (Appendix Table~\ref{tab:loss_ablation_full}); because SAGE and $z$ share an encoder, this gain comes from factorization. Its ARI/NMI is $0.337/0.472$, versus at most $0.009/0.019$ for baselines. On FFHQ, SAGE achieves the best probe accuracy and clustering ($0.983$; $0.939/0.865$), ahead of DINOv3 + SepCLR ($0.976$; $0.915/0.825$). High probe accuracy alone does not ensure well-separated clusters: cVAE and SepVAE reach FFHQ probe accuracies near $0.90$ yet ARI near zero. SAGE helps most when $z$ obscures subtypes.

\subsection{Factor Manipulation}
\label{sec:manipulation}

\begin{wrapfigure}{r}{0.5\linewidth}
\vspace{-8pt}
\centering
\includegraphics[width=\linewidth]{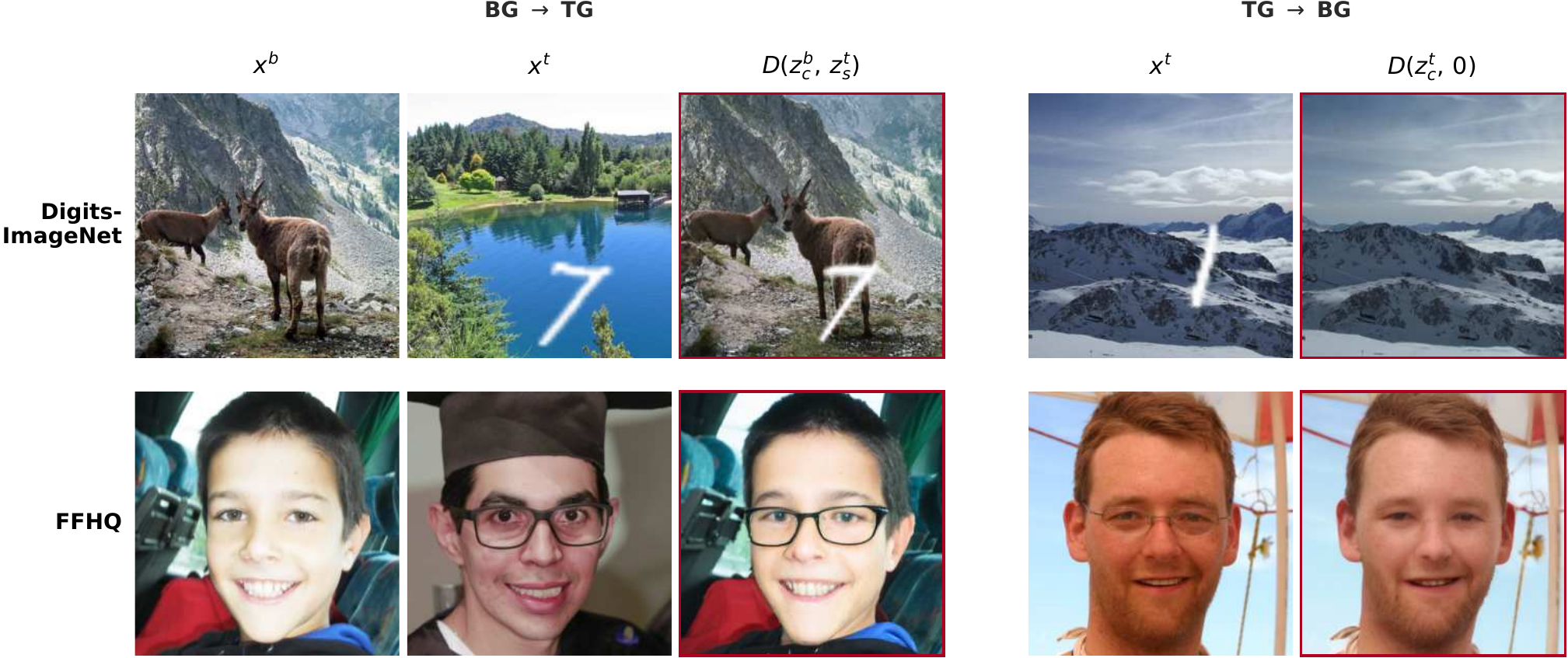}
\caption{\textbf{Salient swapping and erasure.}}
\label{fig:swap-salient}
\vspace{-8pt}
\end{wrapfigure}
\begin{figure}[b]
\centering
\includegraphics[width=0.67\linewidth]{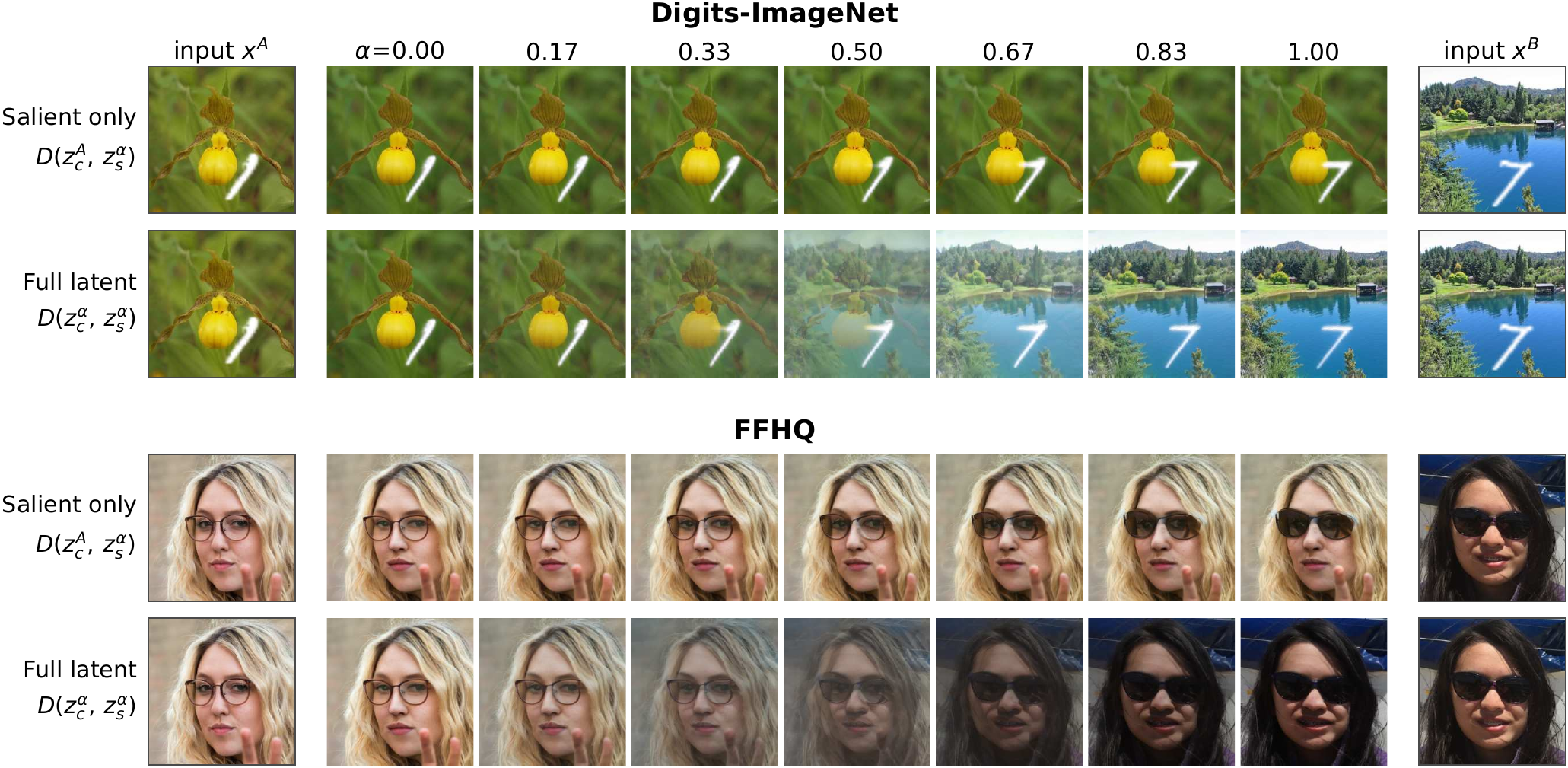}
\caption{\textbf{Salient interpolation.} For two held-out target images $x^A$ and $x^B$, salient-only rows decode $D(z_c^A,z_s^\alpha)$ with $z_s^\alpha=(1-\alpha)z_s^A+\alpha z_s^B$; full-latent rows interpolate both factors.}
\label{fig:interp-salient}
\end{figure}

Because both factors live in the latent space of the frozen decoder, the salient factor can be moved between images, removed, or interpolated, and the result decoded directly. We test whether such edits change only the target attribute. \textbf{Swapping and erasure.} Adding a target image's salient factor to a background image transfers its particular attribute (Fig.~\ref{fig:swap-salient}, left): a digit 7 retains its location and stroke shape, and thick black frames, rather than generic glasses, appear on a new face while the scene and identity remain. Conversely, zeroing a target's salient factor removes the digit or reading glasses while preserving the rest (Fig.~\ref{fig:swap-salient}, right; more in Fig.~\ref{fig:swap-salient-supp}). These edits illustrate that each salient factor carries its own image's target attribute and little else. \textbf{Interpolation.} Interpolating salient factors of two target images while retaining the first common factor changes mainly the target attribute (Fig.~\ref{fig:interp-salient}, salient-only rows): a digit turns from 1 to 7 on a fixed photograph, and clear reading glasses darken into sunglasses on a fixed face. Interpolating both factors cross-fades the whole image (full-latent rows). Salient interpolation thus traverses subtypes while preserving common content.

\subsection{Discovering Disease Subtypes in Retinal OCT}
\label{sec:biomedical}

\begin{figure}[t]
\centering
\includegraphics[width=0.75\linewidth]{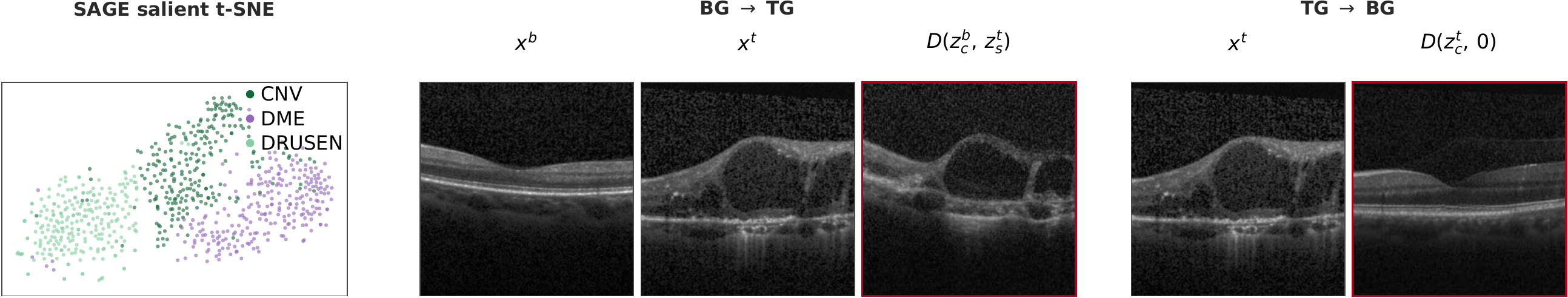}
\caption{\textbf{OCT disease subtypes and salient edits.} Left: salient t-SNE of disease scans, colored by disease subtype labels unused in training. BG$\to$TG adds a disease scan's salient factor to a normal scan; TG$\to$BG removes a disease scan's salient factor.}
\label{fig:oct-extension}
\end{figure}

Unlike digits and eyeglasses, retinal disease subtypes are clinically subtle and vary across patients. On OCT-Kermany, SAGE is trained only with normal/disease labels. \textbf{Disease subtypes.} Its salient space separates the three diseases, with t-SNE groups that overlap mainly at their boundaries (Fig.~\ref{fig:oct-extension}, left), a linear-probe accuracy of $0.960$, and ARI/NMI $0.415/0.415$ (Appendix~\ref{app:oct_stage1_metrics}). \textbf{Disease-specific edits.} Adding a disease scan's salient factor to a normal scan introduces fluid-filled cavities, while removing it yields a retina closer to normal (Fig.~\ref{fig:oct-extension}, middle and right), suggesting disease-related features rather than patient-specific anatomy. These edits are qualitative, not clinical evidence.

\subsection{Salient-Conditioned Generation}
\label{sec:stage2_experiments}
\begin{table}[!t]
\centering
\setlength{\abovecaptionskip}{0pt}
\caption{\textbf{Salient-conditioned generation.} Raw DINOv3 GMP conditions on the unfactorized latent. Subtype Acc.: share of samples assigned the reference subtype by a DINOv3-L classifier trained on real targets. Vendi (diversity) and Cos-to-ref (reference similarity) use 16 samples per reference and DINOv2-L features. \textit{Real images}: classifier accuracy and within-subtype Vendi (Appendix~\ref{app:within_condition_diversity}).}
\label{tab:stage2_generation}
\footnotesize
\setlength{\tabcolsep}{4pt}
\renewcommand{\arraystretch}{1.0}
\resizebox{0.85\linewidth}{!}{%
\begin{tabular}{llccccc}
\toprule
Dataset & Conditioning
& Uncond. gFID\,\textdownarrow
& Cond. gFID\,\textdownarrow
& Subtype Acc.\,\textuparrow
& Vendi\,\textuparrow
& Cos-to-ref\,\textdownarrow \\
\midrule
\multirow{3}{*}{Digits-ImageNet}
& \textit{Real images} & -- & -- & \textit{99.3} & \textit{15.79} & -- \\
& Raw DINOv3 GMP & 8.86 & \textbf{2.40} & 27.70 & 2.65 & 0.794 \\
\rowcolor{gray!15} & \textbf{SAGE Salient GMP} & \textbf{4.76} & 4.09 & \textbf{90.48} & \textbf{12.50} & \textbf{0.116} \\
\midrule
\multirow{3}{*}{FFHQ}
& \textit{Real images} & -- & -- & \textit{99.1} & \textit{10.09} & -- \\
& Raw DINOv3 GMP & 16.07 & \textbf{11.17} & \textbf{98.5} & 2.32 & 0.779 \\
\rowcolor{gray!15} & \textbf{SAGE Salient GMP} & \textbf{15.80} & 16.03 & 96.6 & \textbf{6.96} & \textbf{0.470} \\
\bottomrule
\end{tabular}%
}
\end{table}

Representation-conditioned generation~\citep{li2024return} conditions on a full self-supervised representation; we instead condition on SAGE's pooled salient representation $c_s$. Raw DINOv3 GMP pools the unfactorized latent $z$ identically, testing whether factorization removes reference face or scene information. \textbf{Qualitative results.} Salient conditioning generates new faces and scenes that keep the reference digit or eyewear type, whereas Raw DINOv3 GMP reproduces the reference person or scene almost exactly and often changes the digit (Appendix Fig.~\ref{fig:stage2-qualitative}); Fig.~\ref{fig:stage-2-framework} (top right) shows several samples per reference. \textbf{Quantitative results.} On Digits-ImageNet, salient conditioning raises subtype accuracy from $27.7\%$ to $90.5\%$ and Vendi from $2.65$ to $12.50$ (real: $15.79$), and lowers Cos-to-ref from $0.794$ to $0.116$ (Table~\ref{tab:stage2_generation}). On FFHQ, both conditions preserve eyewear type ($96.6\%$ versus $98.5\%$), but only salient conditioning produces varied samples (Vendi $6.96$ versus $2.32$; Cos-to-ref $0.470$ versus $0.779$). Since the two conditions differ only in the salient encoder, reduced copying supports $z_s$ as a more selective condition. Raw DINOv3 GMP attains lower conditional gFID, consistent with closer reference copying rather than subtype control, while SAGE has lower unconditional gFID on both datasets.

\subsection{Ablation Studies}
\label{sec:ablation}

\begin{table}[!ht]
\centering
\setlength{\abovecaptionskip}{0pt}
\caption{\textbf{Loss ablations on Digits-ImageNet.} Common-only: rFID and SSIM of $D(z_c,\mathbf{0})$ for target images against their digit-free ImageNet backgrounds. Salient probes measure digit identity (target content, higher is better) and ImageNet category (shared content, lower is better). ARI/NMI are computed as in Table~\ref{tab:benchmark_main}; their standard deviations over three $k$-means seeds are at most $0.0007$.}
\label{tab:loss_ablation_main}
\setlength{\tabcolsep}{3pt}
\renewcommand{\arraystretch}{1.15}
\resizebox{\linewidth}{!}{%
\begin{tabular}{l cccc cc cc c}
\toprule
& \multicolumn{4}{c}{Full Reconstruction}
& \multicolumn{2}{c}{Common-only}
& \multicolumn{2}{c}{Salient Linear Probe}
& Clustering \\
\cmidrule(lr){2-5}\cmidrule(lr){6-7}\cmidrule(lr){8-9}\cmidrule(lr){10-10}
Configuration
& rFID\,\textdownarrow & PSNR\,\textuparrow & SSIM\,\textuparrow & LPIPS\,\textdownarrow
& rFID\,\textdownarrow & SSIM\,\textuparrow
& Digit\,\textuparrow & ImageNet\,\textdownarrow
& ARI/NMI\,\textuparrow \\
\midrule
\rowcolor{gray!15}\textbf{SAGE (all objectives)} & 1.78 & 20.56 & 0.571 & 0.216 & 1.89 & 0.569 & 0.950 & 0.045 & 0.337/0.472 \\
w/o swap adversarial ($\mathcal{L}^{G}_{\mathrm{swap}}$) & 1.01 & 20.67 & 0.566 & 0.210 & 4.88 & 0.544 & 0.123 & 0.169 & 0.000/0.001 \\
w/o target sparsity ($\mathcal{L}_{\mathrm{TG\text{-}sp}}$) & 1.15 & 21.06 & 0.598 & 0.193 & 1.36 & 0.585 & 0.923 & 0.302 & 0.274/0.409 \\
w/o Cycle NCE ($\mathcal{L}_{\mathrm{cNCE}}$) & 1.29 & 20.78 & 0.576 & 0.205 & 1.58 & 0.573 & 0.950 & 0.100 & 0.348/0.479 \\
w/o image-space cycle consistency ($\mathcal{L}_{\mathrm{cyc}}$) & 2.84 & 19.88 & 0.562 & 0.236 & 3.46 & 0.559 & 0.951 & 0.086 & 0.360/0.496 \\
w/o latent-space swap consistency ($\mathcal{L}_{\mathrm{lsc}}$) & 1.82 & 20.30 & 0.568 & 0.218 & 2.25 & 0.567 & 0.953 & 0.107 & 0.361/0.494 \\
\bottomrule
\end{tabular}%
}
\end{table}

Table~\ref{tab:loss_ablation_main} removes each objective in turn. The \textbf{swap adversarial loss} is what moves target content into the salient factor: without it, the digit probe drops from $0.950$ to $0.123$ and ARI/NMI to nearly zero, consistent with $z_s$ staying near its zero initialization. The other objectives mainly keep shared content out: removing \textbf{target sparsity}, \textbf{Cycle NCE}, or either \textbf{consistency} loss raises the salient ImageNet probe from $0.045$ to between $0.086$ and $0.302$, and removing the image-space cycle loss also degrades reconstruction (rFID $1.78\to2.84$). We keep all objectives, which together give the lowest salient ImageNet probe accuracy (Appendix~\ref{app:full_ablation}).

\section{Conclusion}
We presented SAGE, which learns salient factors in the high-dimensional spatial latent of a frozen representation autoencoder using only background/target labels. Its factors faithfully reconstruct and edit individual images. Without subtype labels, its salient representation recovers digits, eyewear types, and retinal diseases, reveals finer sunglasses styles and mislabeled images, and conditions the generation of new images that keep a discovered subtype while common content varies.

\subsection*{Reproducibility statement}

We will release code, trained models, and the audited FFHQ labels upon publication. All experiments use public data: MNIST and ImageNet for Digits-ImageNet, FFHQ with the FFHQ Features eyewear annotations, and OCT-Kermany. Appendix~\ref{app:dataset_definitions} describes how each dataset is built, including the Digits-ImageNet compositing, the audited FFHQ evaluation labels, and the OCT preprocessing and splits. Appendices~\ref{app:stage1_details} and~\ref{app:stage2_details} give the Stage~1 and Stage~2 architectures, objectives, optimization settings, and checkpoints, and Appendices~\ref{app:baselines} and~\ref{app:dino_sepclr} give the baseline configurations. Appendices~\ref{app:linear_probe}, \ref{app:ari_protocol}, and~\ref{app:within_condition_diversity} specify the linear-probe, clustering, and generation evaluation protocols. SAGE uses the pretrained RAEv2 encoder and decoder of \citet{singh2026improved}, which remain frozen throughout.

\subsection*{AI use statement}

In this work, we used generative AI to assist with the implementation of experimental methods and code. We have not used generative AI tools for other tasks requiring disclosure, and the remaining required-disclosure tasks are not applicable to this work. Additionally, we used generative AI tools to polish the writing of the manuscript. All AI-assisted experimental code was reviewed and tested by the authors, and the resulting experimental outputs were also verified by the authors. All AI-assisted writing was also reviewed by the authors. We take responsibility for the final content of this work, including text, claims, code and results produced with the aid of generative AI.
\section*{Acknowledgements}

This work was supported by start-up funding awarded to Yuan Yuan by Boston College and by the Boston College Undergraduate Research Fellows program.

\clearpage

\bibliography{iclr2025_conference}
\bibliographystyle{iclr2027_conference}

\typeout{SAGE-SUPPL-COUNTERS: equation=\arabic{equation},figure=\arabic{figure},table=\arabic{table}}
\clearpage
\appendix
\raggedbottom
\makeatletter
\newenvironment{herefigure}{\par\medskip\noindent\begin{minipage}{\linewidth}\def\@captype{figure}}{\end{minipage}\par\medskip}
\makeatother
\section*{Supplementary Materials for SAGE: Salient Factor Discovery and Generation with Visual Foundation Representations}

\section{Additional Experimental Results}
\label{app:additional_experiments}

\subsection{Additional Factor-Decoding Examples}
\label{app:factor_decoding_examples}

\begin{herefigure}
\centering
\includegraphics[width=0.8\linewidth]{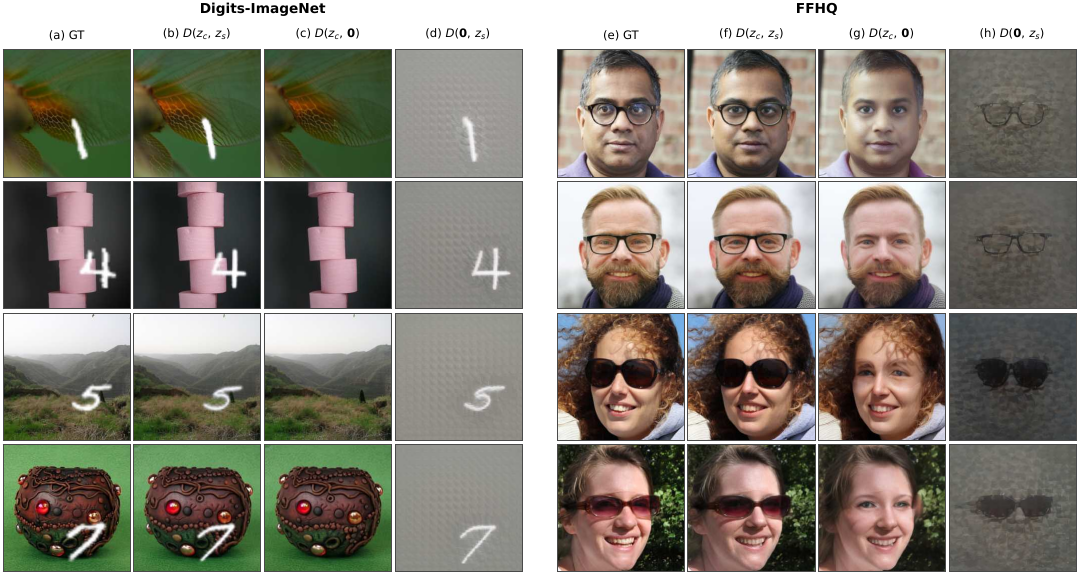}
\caption{\textbf{Additional SAGE factor-decoding examples.} Randomly selected images. For Digits-ImageNet (a--d) and FFHQ (e--h), complete decoding reconstructs the target image, common-only decoding removes the digit or eyeglasses while preserving the scene or identity, and salient-only decoding retains the digit's shape and location or the specific eyewear style with little common content.}
\label{fig:factor-decoding-supp}
\end{herefigure}

Figure~\ref{fig:factor-decoding-supp} complements the cross-method comparison in Fig.~\ref{fig:factor-decoding-benchmark} with SAGE-only examples. It shows that $D(z_c,\mathbf{0})$ removes the target attribute while retaining common content, whereas $D(\mathbf{0},z_s)$ preserves image-specific target details rather than a generic attribute.

\subsection{Clustering across Representation Spaces}
\label{app:ari_spaces}
Table~\ref{tab:ari_spaces} compares $k$-means on each method's $\ell_2$-normalized salient vectors with $k$-means after PCA projection, using the protocol in Appendix~\ref{app:ari_protocol}.

\begin{table}[!htb]
\centering
\caption{\textbf{Clustering readout ablation on $\ell_2$-normalized salient representations (ARI/NMI).}
All vectors are normalized before optional projection. Raw denotes no dimensionality reduction; PCA-50 retains $\min(50,\mathrm{Dim.})$ components. Scores are means over $k$-means seeds $\{0,1,2\}$. FFHQ uses manually verified annotations, with 39 no-glasses images excluded for $k=2$ and included for $k=3$.}
\label{tab:ari_spaces}
\vspace{6pt}
\scriptsize
\setlength{\tabcolsep}{3.5pt}
\renewcommand{\arraystretch}{1.15}
\begin{tabular}{l c ccc ccc ccc}
\toprule
& & \multicolumn{3}{c}{\makecell{Digits-ImageNet\\($k=10$, $n=24{,}739$)}}
& \multicolumn{3}{c}{\makecell{FFHQ\\($k=2$, $n=3{,}398$)}}
& \multicolumn{3}{c}{\makecell{FFHQ\\($k=3$, $n=3{,}437$)}} \\
\cmidrule(lr){3-5}\cmidrule(lr){6-8}\cmidrule(lr){9-11}
Method & Dim. & PCA-2D & PCA-50 & Raw & PCA-2D & PCA-50 & Raw & PCA-2D & PCA-50 & Raw \\
\midrule
\multicolumn{11}{l}{\textit{ARI}\,\textuparrow} \\
cVAE & 64 & 0.000 & 0.001 & 0.001 & 0.001 & 0.000 & 0.001 & 0.036 & 0.017 & 0.016 \\
SepVAE & 64 & 0.006 & 0.009 & 0.009 & -0.001 & -0.001 & -0.001 & 0.012 & 0.019 & 0.017 \\
Double-InfoGAN & 64 & 0.005 & 0.001 & 0.001 & 0.324 & 0.292 & 0.298 & 0.247 & 0.303 & 0.303 \\
SepCLR & 32 & 0.000 & 0.000 & 0.000 & 0.808 & 0.822 & 0.822 & 0.441 & 0.455 & 0.455 \\
DINOv3 + SepCLR & 32 & 0.000 & 0.000 & 0.000 & 0.415 & 0.915 & 0.915 & 0.187 & 0.453 & 0.453 \\
\midrule
\rowcolor{gray!15}\textbf{SAGE (Ours)} & 1024 & \textbf{0.187} & \textbf{0.336} & \textbf{0.337} & \textbf{0.930} & \textbf{0.937} & \textbf{0.939} & \textbf{0.902} & \textbf{0.936} & \textbf{0.937} \\
\midrule
\multicolumn{11}{l}{\textit{NMI}\,\textuparrow} \\
cVAE & 64 & 0.001 & 0.003 & 0.003 & 0.001 & 0.001 & 0.001 & 0.047 & 0.026 & 0.025 \\
SepVAE & 64 & 0.012 & 0.019 & 0.019 & 0.000 & 0.000 & 0.000 & 0.017 & 0.031 & 0.029 \\
Double-InfoGAN & 64 & 0.010 & 0.003 & 0.003 & 0.318 & 0.297 & 0.299 & 0.319 & 0.380 & 0.380 \\
SepCLR & 32 & 0.001 & 0.001 & 0.001 & 0.680 & 0.696 & 0.696 & 0.522 & 0.534 & 0.534 \\
DINOv3 + SepCLR & 32 & 0.001 & 0.001 & 0.001 & 0.379 & 0.825 & 0.825 & 0.288 & 0.550 & 0.550 \\
\midrule
\rowcolor{gray!15}\textbf{SAGE (Ours)} & 1024 & \textbf{0.332} & \textbf{0.471} & \textbf{0.472} & \textbf{0.850} & \textbf{0.863} & \textbf{0.865} & \textbf{0.813} & \textbf{0.864} & \textbf{0.865} \\
\bottomrule
\end{tabular}
\end{table}

SAGE achieves the highest ARI and NMI in the normalized original space and both PCA readouts across the settings in Table~\ref{tab:ari_spaces}. On FFHQ, its original-space scores are 0.939/0.865 for $k=2$ and 0.937/0.865 for $k=3$, compared with 0.915/0.825 and 0.453/0.550 for DINOv3 + SepCLR.

\subsection{Salient Representation Visualizations Across Methods}
\label{app:tsne_baselines}
\begin{figure}[!htb]
\centering
\includegraphics[width=0.8\linewidth]{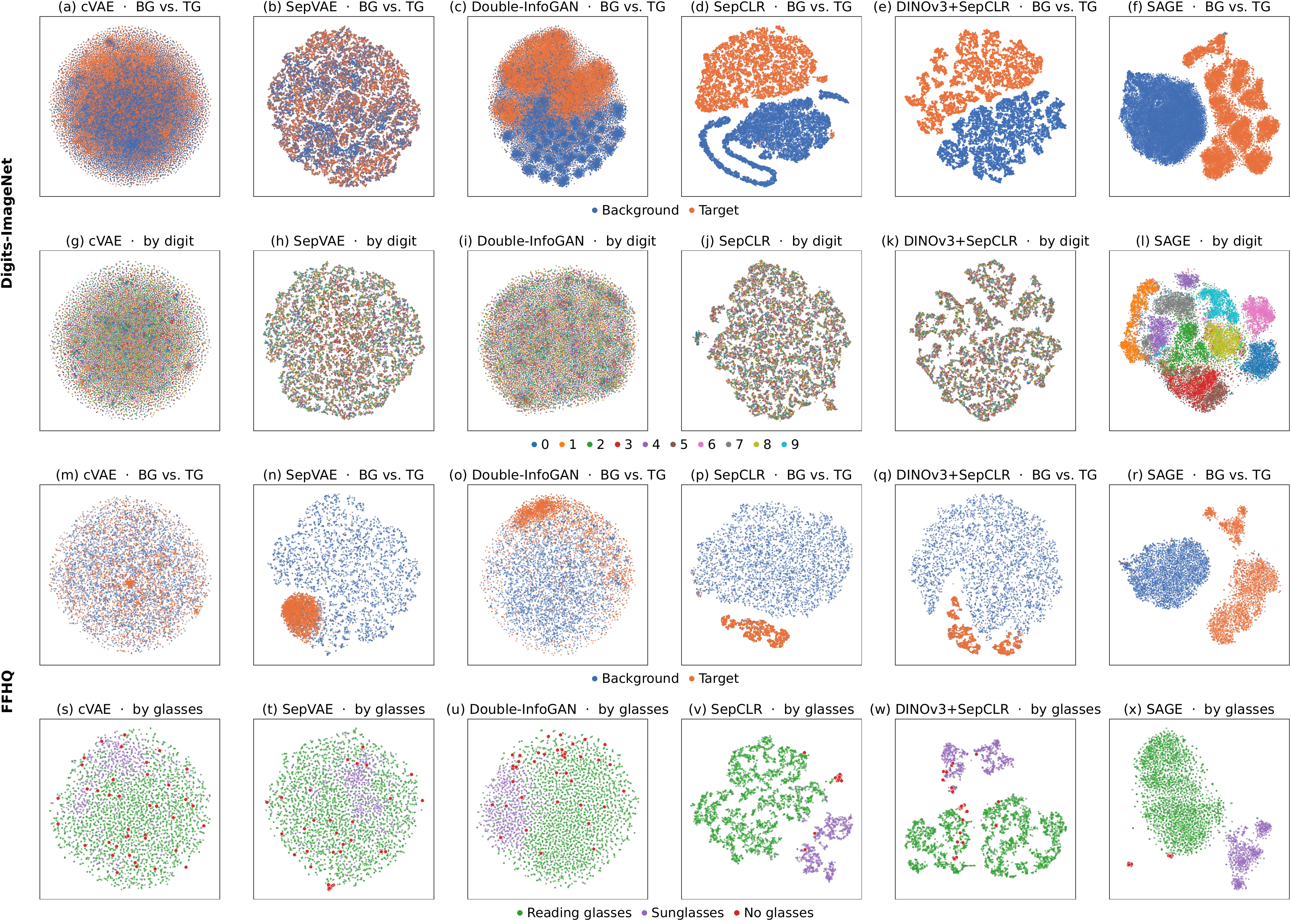}
\caption{\textbf{t-SNE visualizations of salient representations across contrastive analysis methods.}
Columns show cVAE, SepVAE, Double-InfoGAN, SepCLR, DINOv3 + SepCLR, and SAGE. Rows show Digits-ImageNet background-versus-target and digit-colored target representations, then their FFHQ counterparts with audited eyewear labels. Blue/orange denote background/target; green/purple/red denote reading glasses/sunglasses/audited no-glasses images.}
\label{fig:tsne_baselines}
\end{figure}

Figure~\ref{fig:tsne_baselines} compares the salient representations of all methods. On Digits-ImageNet, SepCLR and DINOv3 + SepCLR separate background from target but mix digit identities, whereas SAGE forms clear digit groups. On FFHQ, SepCLR, DINOv3 + SepCLR, and SAGE all reveal eyewear-related groups. Table~\ref{tab:silhouette_scores} quantifies the background/target separation: SAGE has the highest silhouette score in both the t-SNE and original spaces on both datasets.

\begin{table}[!htb]
\centering
\caption{\textbf{Silhouette scores for BG-vs.-TG salient separation in t-SNE and original spaces.} Scores treat background (BG) and target (TG) salient representations as the two groups. The silhouette coefficient~\citep{rousseeuw1987silhouettes} measures within-group cohesion relative to separation; higher is better.}
\label{tab:silhouette_scores}
\vspace{6pt}
\scriptsize
\setlength{\tabcolsep}{4pt}
\renewcommand{\arraystretch}{1.15}
\begin{tabular}{lcccc}
\toprule
Method & Digits t-SNE & Digits original & FFHQ t-SNE & FFHQ original \\
\midrule
cVAE & 0.022 & 0.006 & 0.004 & 0.027 \\
SepVAE & 0.003 & 0.006 & 0.394 & 0.206 \\
Double-InfoGAN & 0.248 & 0.013 & 0.188 & 0.050 \\
SepCLR & 0.381 & 0.546 & 0.449 & 0.752 \\
DINOv3 + SepCLR & 0.365 & 0.642 & 0.387 & 0.551 \\
\rowcolor{gray!15}\textbf{SAGE} & \textbf{0.417} & \textbf{0.820} & \textbf{0.462} & \textbf{0.873} \\
\bottomrule
\end{tabular}
\end{table}

\subsection{Fine-Grained Subtype Structure within Sunglasses}
\label{app:sunglasses_subtypes}

\begin{figure}[!htb]
\centering
\includegraphics[width=0.7\linewidth]{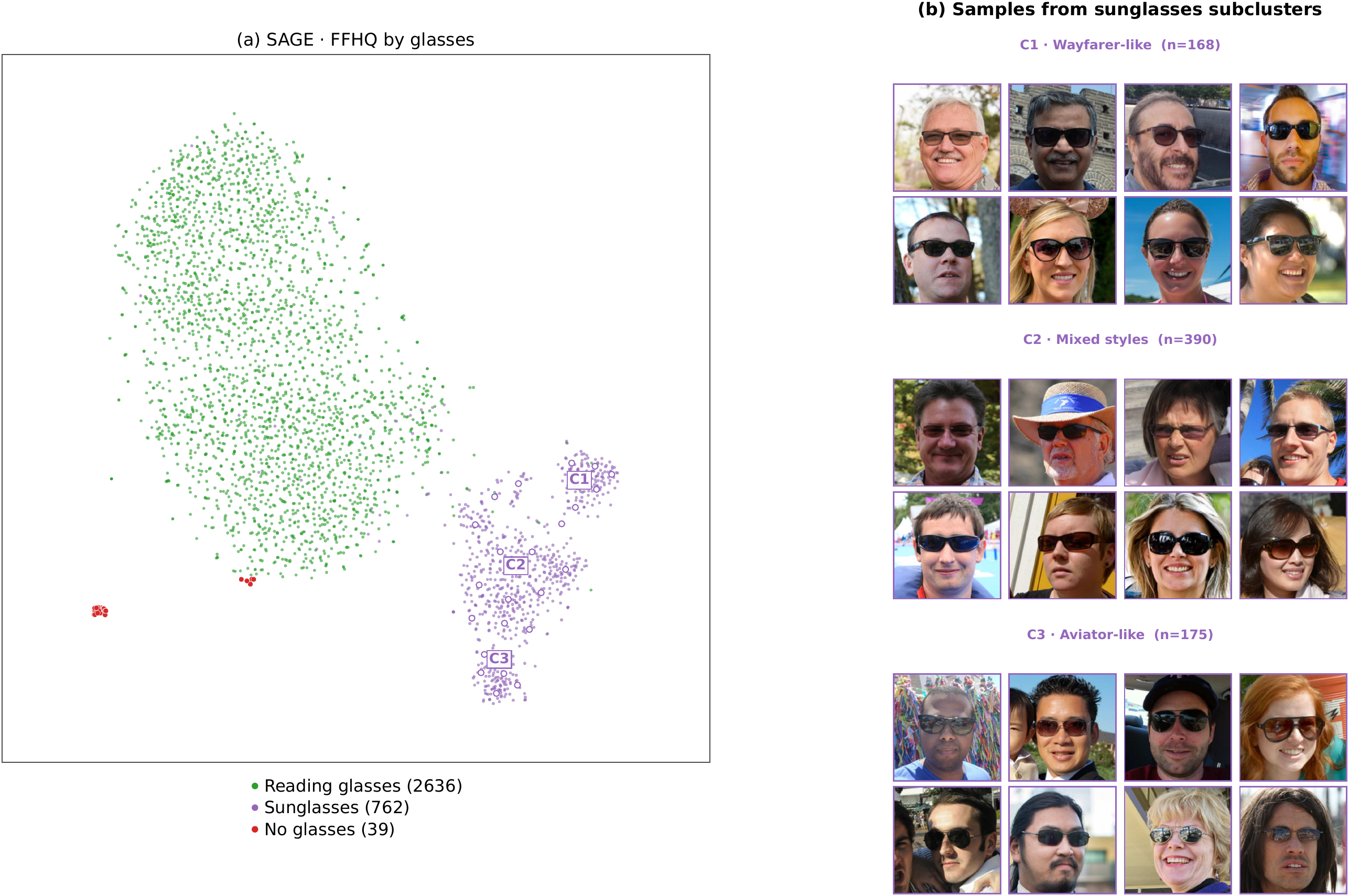}
\caption{\textbf{Fine-grained sunglasses structure in SAGE's salient representation.}
Left: FFHQ target embeddings colored by audited eyewear labels, with C1--C3 marking subgroups within sunglasses. Right: example images from each subgroup: Wayfarer-like (C1, $n=168$), mixed styles (C2, $n=390$), and Aviator-like (C3, $n=175$). Group names describe the displayed examples; no style labels are used in training.}
\label{fig:sunglasses-subtypes}
\end{figure}

Figure~\ref{fig:sunglasses-subtypes} looks inside the sunglasses subtype. Without labels for finer styles, SAGE's salient representation splits sunglasses into three groups, from a compact C1 to a broader C2 and a smaller C3 at its lower end: C1 (Wayfarer-like) collects thick frames with angular or rectangular lenses; C2 (mixed styles) spans rounded, narrow, and sporty frames with varied lens tint; and C3 (Aviator-like) collects thin-rimmed, rounded or teardrop-shaped lenses, often reflective. The three groups share one coarse label yet differ in frame geometry and lens shape, showing that the salient factor captures appearance structure finer than the annotated subtypes.

\subsection{FFHQ Eyewear Label Auditing and Remaining SAGE Errors}
\label{app:azure_label_errors}

\begin{figure}[t]
\centering
\includegraphics[width=0.85\linewidth]{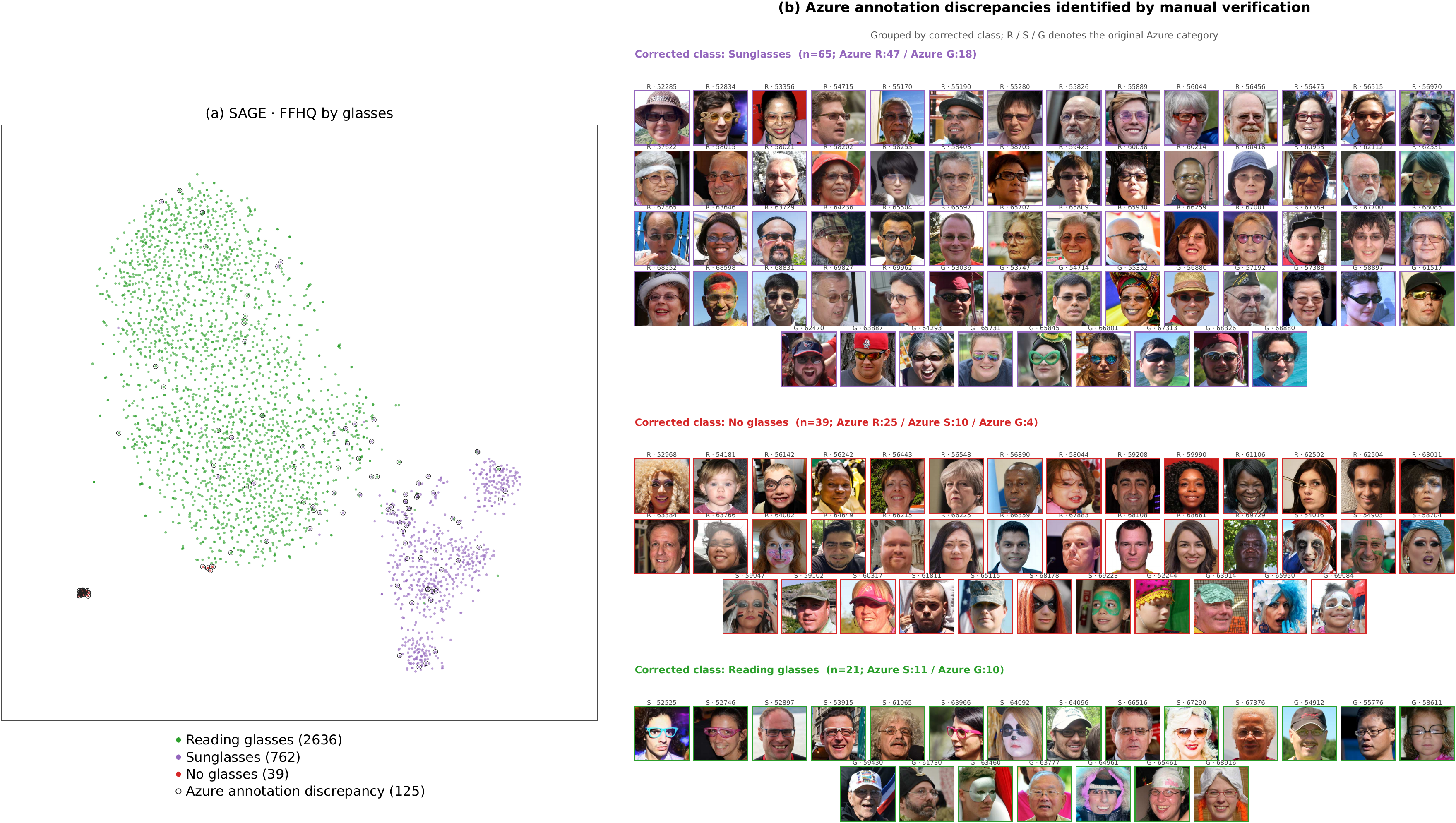}
\caption{\textbf{Azure eyewear annotation errors identified by manual inspection of FFHQ.} Left: t-SNE of SAGE's target salient representations, colored by manually verified labels; outlined points mark the 125 Azure label disagreements. Right: all 125 corrected images, grouped by verified label: sunglasses (65), no glasses (39), and reading glasses (21). Tile titles give the original Azure category and image ID.}
\label{fig:azure_label_errors}
\end{figure}

In the t-SNE of SAGE's salient representations of FFHQ target images, one group stood apart (Section~\ref{sec:stage1_experiments}). Inspecting its images showed faces without eyeglasses that the Azure Face API had labeled as wearing glasses. We therefore manually audited all 3,437 target images in our FFHQ validation split and corrected 125 Azure Face API subtype labels (3.64\%). We then examined the locations and neighborhoods of all manually verified samples in SAGE's salient representation. Some faces with face paint or hat-brim shadows were incorrectly labeled as sunglasses. The final class counts are reported in Appendix~\ref{app:dataset_definitions}. Figure~\ref{fig:azure_label_errors} shows all 125 corrected samples alongside their locations in the salient-space visualization.

Annotators assigned each label from the visible eyewear in the image, without seeing SAGE's cluster assignments or t-SNE. Table~\ref{tab:benchmark_main} uses the same corrected labels for all methods. SAGE surfaced the no-glasses errors, 34 of which form a separate group; corrections between reading glasses and sunglasses were identified by the manual audit rather than by SAGE.

\begin{figure}[!ht]
\centering
\includegraphics[width=0.85\linewidth]{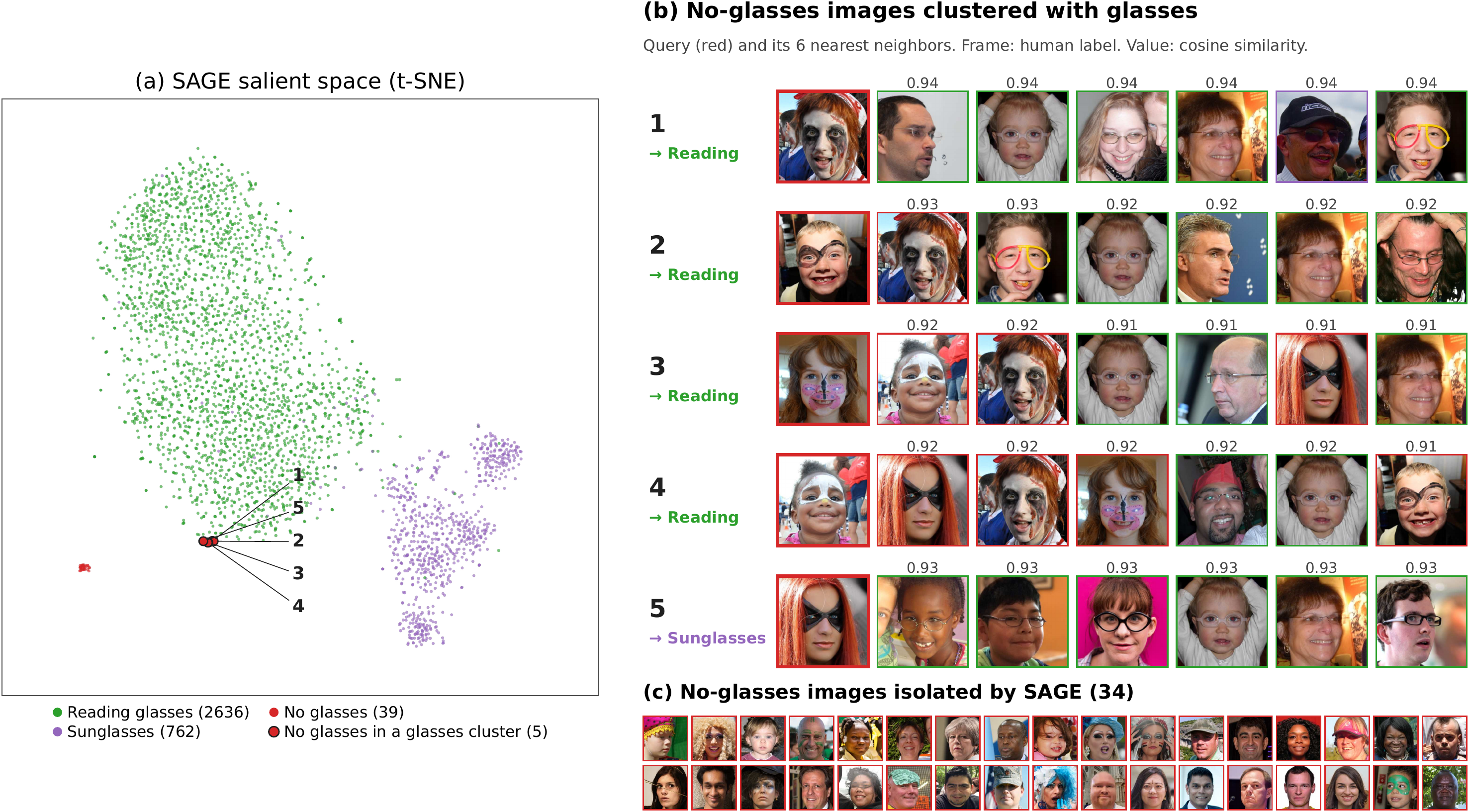}
\caption{\textbf{Remaining SAGE errors among manually verified no-glasses FFHQ images.}
(a) Target salient-space t-SNE, colored by audited labels: reading glasses (green, 2,636), sunglasses (purple, 762), and no glasses (red, 39). Black-outlined red points identify five no-glasses cases associated with glasses groups. (b) Each numbered row shows one query and its six nearest neighbors; border colors indicate human labels and values report cosine similarity. Arrows mark the group each query is associated with. (c) The other 34 no-glasses images form a separate group.}
\label{fig:ffhq-noglasses-failures}
\end{figure}

\paragraph{Remaining no-glasses cases.}
Figure~\ref{fig:ffhq-noglasses-failures} examines the 39 audited no-glasses images in the target dataset. SAGE places 34 of them in a separate group (panel c); the other five lie with glasses groups, four with reading glasses and one with sunglasses (panel b). These five show face paint around the eyes or a dark costume eye mask, whose contours resemble eyewear, and their nearest neighbors include both genuine glasses and similarly decorated no-glasses faces (cosine similarity $0.91$--$0.94$). The FFHQ scores in Table~\ref{tab:benchmark_main} use only the verified reading-glasses and sunglasses images, so these cases do not enter them.

\subsection{Additional Swapping and Erasure Examples}
\label{app:swap_examples}

\begin{figure}[t]
\centering
\includegraphics[width=0.7\linewidth]{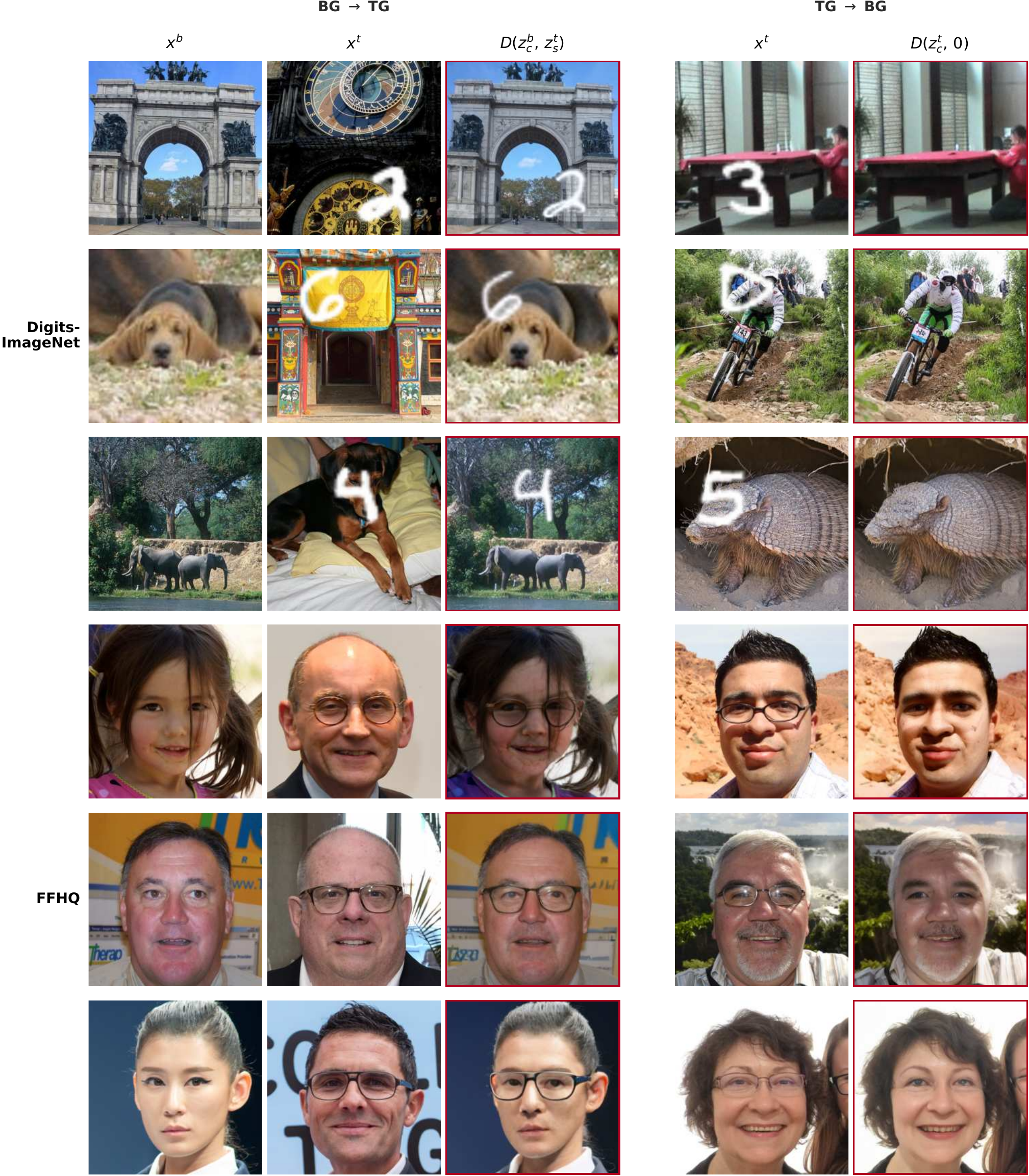}
\caption{\textbf{Additional salient swapping and erasure examples.} Left (BG $\rightarrow$ TG): $D(z_c^b,z_s^t)$ combines a background image's common factor with a target image's salient factor. Right (TG $\rightarrow$ BG): $D(z_c^t,\mathbf{0})$ decodes the target common factor alone. Rows show Digits-ImageNet and FFHQ examples; red borders mark decoded outputs.}
\label{fig:swap-salient-supp}
\end{figure}

Figure~\ref{fig:swap-salient-supp} extends Fig.~\ref{fig:swap-salient} with more examples of both manipulations on Digits-ImageNet and FFHQ. Swapping a target salient factor into a background common factor transfers the target attribute to the donor content, whereas decoding a target common factor alone removes that attribute.

\subsection{Stage 1 Metrics on OCT-Kermany}
\label{app:oct_stage1_metrics}
To assess disease-subtype structure under normal/disease labels (Section~\ref{sec:biomedical}), Table~\ref{tab:oct_stage1} reports salient linear-probe and clustering metrics on the official OCT-Kermany test set~\citep{kermany2018identifying}, alongside reconstruction quality. Appendix~\ref{app:oct_kermany_details} describes the dataset, splits, and preprocessing, and Appendix~\ref{app:stage1_training} the training settings.

\begin{table}[!htb]
\centering
\caption{\textbf{Stage-1 metrics on OCT-Kermany.}
Evaluation uses the official test set~\citep{kermany2018identifying}, with 250 normal images and 250 images from each of CNV, DME, and DRUSEN, from patients not included in training or validation. LP Acc. reports mean stratified five-fold linear-probe accuracy on the 750 target images; ARI/NMI use $k$-means directly on SAGE's $\ell_2$-normalized, 1,024-dimensional salient representations obtained by global max pooling (GMP), without dimensionality reduction. With only 1,000 evaluation images, rFID is sensitive to finite-sample bias and is not directly comparable to scores computed on larger evaluation sets.}
\label{tab:oct_stage1}
\vspace{6pt}
\scriptsize
\setlength{\tabcolsep}{4pt}
\renewcommand{\arraystretch}{1.15}
\begin{tabular}{l cccc cc}
\toprule
Method
& rFID\,\textdownarrow
& PSNR\,\textuparrow
& SSIM\,\textuparrow
& LPIPS\,\textdownarrow
& LP Acc.\,\textuparrow
& ARI/NMI\,\textuparrow \\
\midrule
\rowcolor{gray!15}SAGE (Ours)
& 27.21 & 24.55 & 0.448 & 0.273 & 0.960 & 0.415/0.415 \\
\bottomrule
\end{tabular}
\end{table}

\subsection{Salient-Conditioned Generation}

\begin{figure}[t]
\centering
\includegraphics[width=0.85\linewidth]{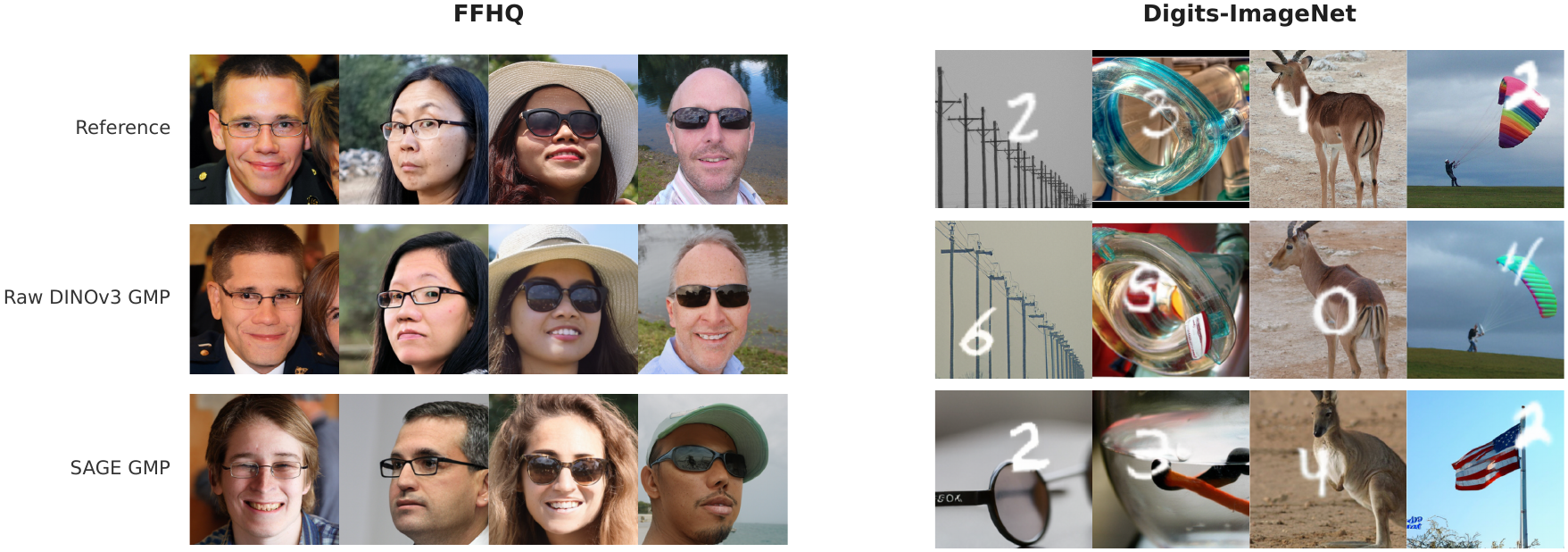}
\caption{\textbf{Generation conditioned on a reference image.} For FFHQ (left) and Digits-ImageNet (right), each column shows a reference (top) and one sample conditioned on Raw DINOv3 GMP (middle) or on SAGE's salient representation (bottom).}
\label{fig:stage2-qualitative}
\end{figure}

\paragraph{Qualitative results.}
Salient conditioning generates new faces and scenes that preserve the reference digit or eyewear type (Fig.~\ref{fig:stage2-qualitative}), whereas Raw DINOv3 GMP reproduces the reference person or scene almost exactly and may change or distort the digit. Multiple samples from one reference, as measured by Vendi score, appear in Fig.~\ref{fig:stage-2-framework} (top right).

\subsection{Full Loss Ablation}
\label{app:full_ablation}
Table~\ref{tab:loss_ablation_full} expands the ablations in Section~\ref{sec:ablation} with the unfactorized latent as a reference and probes of both factors.
\begin{table}[tb]
\centering
\caption{\textbf{Full loss ablation of SAGE on Digits-ImageNet.} ARI/NMI use Euclidean $k$-means on $\ell_2$-normalized original salient representations, without dimensionality reduction (Appendix~\ref{app:ari_protocol}). ARI/NMI are means over $k$-means seeds $\{0,1,2\}$; standard deviations are omitted because they are at most $0.0007$. Digit/ImageNet LP: linear-probe accuracy for digit or ImageNet labels from the indicated representation; Raw $z$: the unfactorized latent, whose probes use $z$ itself. Objectives are named as in Table~\ref{tab:loss_ablation_main}.}
\label{tab:loss_ablation_full}
\vspace{6pt}
\scriptsize
\setlength{\tabcolsep}{3pt}
\renewcommand{\arraystretch}{1.15}
\begin{tabular}{l cccc cc cc cc c}
\toprule
& \multicolumn{4}{c}{Full reconstruction} & \multicolumn{2}{c}{Common-only} & \multicolumn{2}{c}{Digit LP} & \multicolumn{2}{c}{ImageNet LP} & Clustering \\
\cmidrule(lr){2-5}\cmidrule(lr){6-7}\cmidrule(lr){8-9}\cmidrule(lr){10-11}\cmidrule(lr){12-12}
Configuration & rFID\,\textdownarrow & PSNR\,\textuparrow & SSIM\,\textuparrow & LPIPS\,\textdownarrow & rFID\,\textdownarrow & SSIM\,\textuparrow & $z_s$\,\textuparrow & $z_c$\,\textdownarrow & $z_c$\,\textuparrow & $z_s$\,\textdownarrow & ARI/NMI\,\textuparrow \\
\midrule
Raw $z$ & 0.39 & 22.18 & 0.613 & 0.157 & N/A & N/A & 0.328 & N/A & 0.765 & N/A & 0.000/0.001 \\
\midrule
\rowcolor{gray!15}\textbf{SAGE (all)} & 1.78 & 20.56 & 0.571 & 0.216 & 1.89 & 0.569 & 0.950 & 0.336 & 0.765 & 0.045 & 0.337/0.472 \\
\midrule
w/o $\mathcal{L}^{G}_{\mathrm{swap}}$ & 1.01 & 20.67 & 0.566 & 0.210 & 4.88 & 0.544 & 0.123 & 0.356 & 0.763 & 0.169 & 0.000/0.001 \\
w/o $\mathcal{L}_{\mathrm{TG\text{-}sp}}$ & 1.15 & 21.06 & 0.598 & 0.193 & 1.36 & 0.585 & 0.923 & 0.656 & 0.758 & 0.302 & 0.274/0.409 \\
w/o $\mathcal{L}_{\mathrm{cNCE}}$ & 1.29 & 20.78 & 0.576 & 0.205 & 1.58 & 0.573 & 0.950 & 0.335 & 0.767 & 0.100 & 0.348/0.479 \\
$\mathcal{L}_{\mathrm{BG\text{-}sp}}$: $\ell_2\to\ell_1$ & 1.74 & 20.17 & 0.568 & 0.218 & 2.00 & 0.566 & 0.953 & 0.320 & 0.764 & 0.067 & 0.379/0.505 \\
w/o $\mathcal{L}_{\mathrm{cyc}}$ & 2.84 & 19.88 & 0.562 & 0.236 & 3.46 & 0.559 & 0.951 & 0.366 & 0.762 & 0.086 & 0.360/0.496 \\
w/o $\mathcal{L}_{\mathrm{lsc}}$ & 1.82 & 20.30 & 0.568 & 0.218 & 2.25 & 0.567 & 0.953 & 0.353 & 0.761 & 0.107 & 0.361/0.494 \\
w/o $\mathcal{L}_{\mathrm{BG\text{-}sp}}$ & 1.89 & 20.44 & 0.573 & 0.218 & 2.26 & 0.571 & 0.951 & 0.305 & 0.760 & 0.061 & 0.356/0.487 \\
\bottomrule
\end{tabular}
\end{table}
\paragraph{Cycle NCE and shared-category leakage.}
Cycle NCE keeps shared ImageNet content out of the salient factor: it lowers the salient ImageNet probe from $0.100$ to $0.045$ while digit accuracy stays at $0.950$.

\paragraph{Choosing the background penalty.}
We use a squared $\ell_2$ background penalty rather than $\ell_1$ because it better keeps shared content out of the salient factor (salient ImageNet probe $0.045$ vs.\ $0.067$) and improves common-only reconstruction (rFID $1.89$ vs.\ $2.00$).

\section{Datasets, Baselines, and Evaluation Details}
\label{app:evaluation_details}

\subsection{Dataset Details}
\label{app:dataset_definitions}
For all datasets, training uses only background/target labels. Subtype labels are used only for evaluation and are not provided during representation learning.

\paragraph{Digits-ImageNet.}
\label{app:digits_imagenet}
We construct the target dataset by superimposing MNIST digits on ImageNet photographs; the background dataset contains photographs without added digits. We retain the ImageNet training and validation splits and randomly assign approximately half of each split to each dataset, using different random seeds for the training and validation splits. Background and target datasets use disjoint source photographs. All available ImageNet class directories are included, with no explicit class exclusions.

Photographs are center-cropped to a square and resized to $512\times512$ pixels. Each target receives one MNIST example sampled uniformly with replacement. Training images use the MNIST training set, while validation images use the MNIST test set. Digits are resized to $224\times224$ pixels using bilinear interpolation and rendered in white. The top-left coordinates are sampled uniformly from $\{0,\ldots,288\}$ along each axis, keeping the digit patch within the image. We use the resized digit intensities as an alpha mask with an opacity multiplier of 1.0. Digit size and color are fixed, and digit classes are not explicitly balanced.

Linear-probe and clustering evaluations use 24,739 target validation images. The per-class counts for digits 0--9 are 2,464, 2,699, 2,685, 2,521, 2,358, 2,199, 2,337, 2,613, 2,400, and 2,463.

\paragraph{FFHQ.}
\label{app:ffhq_subtypes}
The original FFHQ dataset does not provide attribute-level labels indicating whether a face wears glasses. We use the eyewear annotations from the FFHQ Features Dataset~\citep{ffhqfeatures}, which were generated automatically using Microsoft's Azure Face API, to construct the background and target datasets. During training, we use 10,000 images without glasses as the background dataset and 10,000 images with glasses as the target dataset. Because the annotations are automatic, they can contain errors. After our manual audit of all 3,437 validation images (Appendix~\ref{app:azure_label_errors}), the validation set contains 2,636 images with reading glasses, 762 with sunglasses, and 39 without glasses.

\paragraph{OCT-Kermany.}
\label{app:oct_kermany_details}
We use the OCT-Kermany data of \citet{kermany2018identifying}, treating normal scans as background and CNV, DME, and DRUSEN scans jointly as target. The original Kermany export contains saturated-white borders introduced by acquisition tilt or registration correction, rather than by our dataset construction. These borders can extend 300 pixels into a 496-pixel-high scan. Before cleaning, 60.1\% of the 109,309 scans have at least one edge for which more than half the pixels are saturated white, creating an easily exploitable non-anatomical cue for an encoder.

We preprocess scans in grayscale. First, we remove rows or columns that are at least 95\% white (intensity $\geq 0.90$) along their full extent. To preserve retinal tissue under oblique white wedges, we then replace only near-white regions connected to an image boundary and larger than 64 pixels, rather than rectangularly cropping to the deepest wedge. Replacement pixels are Gaussian noise matched to the mean and standard deviation of the darkest 40\% of pixels in the scan, which correspond to the vitreous. Scans for which this replacement would cover over 60\% of the image are retained unchanged. Finally, we preserve aspect ratio, bicubically resize to height 384, and center-crop to $384\times384$ before storing RGB images. This avoids the aspect-ratio distortions caused by directly resizing the eight native scan widths to a square. During representation learning, images are resized from $384$ to $256$ for DINOv3, without data augmentation.

Table~\ref{tab:oct_splits} gives the resulting splits. We train on the cleaned training set (97,478 scans). All OCT evaluations use the official 1,000-image test set, whose patients do not appear in training.

\begin{table}[!htb]
\centering
\caption{\textbf{OCT-Kermany split statistics after preprocessing.} CNV, DME, and DRUSEN are the target diseases; normal scans are background.}
\label{tab:oct_splits}
\vspace{6pt}
\scriptsize
\setlength{\tabcolsep}{4pt}
\renewcommand{\arraystretch}{1.15}
\begin{tabular}{lrrrrrr}
\toprule
Split & NORMAL & CNV & DME & DRUSEN & Total & Patients \\
\midrule
Train & 46,026 & 33,485 & 10,213 & 7,754 & 97,478 & 4,772 \\
Val. & 5,114 & 3,720 & 1,135 & 862 & 10,831 & 3,329 \\
Test & 250 & 250 & 250 & 250 & 1,000 & 635 \\
\bottomrule
\end{tabular}
\end{table}

\subsection{Baseline Details}
\label{app:baselines}
We use the default configurations of the generative CA baselines cVAE~\citep{abid2019contrastive}, SepVAE~\citep{louiset2023sepvae}, and Double-InfoGAN~\citep{carton2024double}, and of the contrastive CA baseline SepCLR~\citep{louiset2024separating}. Double-InfoGAN uses its native $128\times128$ input resolution; all remaining baselines use $256\times256$ inputs. DINOv3 + SepCLR trains SepCLR's objectives on SAGE's frozen RAEv2 encoder with the same common/salient encoder architecture (Appendix~\ref{app:dino_sepclr}), isolating the contribution of the training framework from that of the visual backbone. We additionally evaluate the unfactorized RAEv2 latent $z$ (DINOv3 with MLS), denoted DINOv3 in figures. For generation, Raw DINOv3 GMP pools $z$ identically to salient conditioning, so the two conditions differ only in whether the salient encoder is applied.

\subsection{DINOv3 + SepCLR Baseline}
\label{app:dino_sepclr}
DINOv3 + SepCLR shares SAGE's frozen RAEv2 encoder (DINOv3-L with multi-layer summation and RAEv2 latent normalization) and common/salient encoder architecture, and differs in its objectives and evaluation readout (Table~\ref{tab:dino_sepclr_components}). Each branch applies global average pooling (GAP), following SepCLR, to obtain a 1,024-dimensional vector. A fully connected layer maps it to the 32-dimensional representation used for evaluation. A separate MLP projector ($32\!\to\!128\!\to\!\mathrm{BN}\!\to\!\mathrm{ReLU}\!\to\!32$, with batch normalization, BN) supplies only the contrastive and $k$-JEM losses; its output is not used for evaluation. Only the two common/salient Transformers, two fully connected layers, and two projectors are trained. We retain SepCLR's alignment, uniformity, salient regularization, and JEM losses and its original loss hyperparameters~\citep{louiset2024separating}.

\begin{table}[!htb]
\centering
\caption{\textbf{Components of DINOv3 + SepCLR and SAGE.}}
\label{tab:dino_sepclr_components}
\vspace{6pt}
\scriptsize
\setlength{\tabcolsep}{4pt}
\renewcommand{\arraystretch}{1.15}
\begin{tabular}{lll}
\toprule
Component & DINOv3 + SepCLR & SAGE \\
\midrule
Frozen visual backbone & DINOv3-L (MLS) & DINOv3-L (MLS) \\
Latent normalization & RAEv2 & RAEv2 \\
Common/salient Transformers & Shared architecture & Shared architecture \\
Evaluation pooling & GAP & GMP \\
Evaluation representation & 32 dimensions & 1,024 dimensions \\
Training objectives & SepCLR objectives & SAGE objectives \\
\bottomrule
\end{tabular}
\end{table}

\subsection{Linear-Probe Evaluation}
\label{app:linear_probe}
We use the evaluation pools defined in Appendix~\ref{app:dataset_definitions}.

\paragraph{Digits-ImageNet.}
\label{app:digits_probe}
We evaluate frozen representations with stratified five-fold cross-validation. We standardize the features and evaluate them using a logistic-regression classifier. No class weighting is used, and predictions are scored using ordinary accuracy, as the digit classes are approximately balanced. Reported digit linear-probe scores are mean accuracies across the five folds.

\paragraph{FFHQ.}
\label{app:ffhq_probe}
We evaluate frozen salient representations on all 3,398 images with the manually verified reading-glasses/sunglasses labels, using class-balanced logistic regression and stratified five-fold cross-validation. The score is balanced accuracy, the mean of the two class recalls. Table~\ref{tab:benchmark_main} reports the mean across folds. The main FFHQ SAGE result uses target salient sparsity weight $\lambda=0.01$.

\subsection{Clustering and t-SNE Protocol}
\label{app:ari_protocol}

\paragraph{Salient representation.}
Clustering uses only target-group salient representations, extracted with the trained encoders frozen. The baseline representations used for evaluation are flat vectors: 32 dimensions for SepCLR and DINOv3 + SepCLR and 64 for cVAE, SepVAE, and Double-InfoGAN. SAGE produces an $N\times1024$ salient token representation, which we reduce to a 1,024-dimensional vector by channel-wise global max pooling (GMP) over the token axis, as in linear-probe evaluation. These pooled SAGE vectors define SAGE's original salient representation space. Within each quantitative evaluation setting, all methods use the same images and labels.

\paragraph{Cluster count.}
We use the evaluation pools defined in Appendix~\ref{app:dataset_definitions}. For all methods, the number of clusters is set to the number of evaluation classes: $k=10$ for Digits-ImageNet, and $k=2$ or $k=3$ for the two FFHQ settings with audited labels, and $k=3$ for the three OCT-Kermany diseases. Table~\ref{tab:benchmark_main} uses the FFHQ two-class setting, while Table~\ref{tab:ari_spaces} reports both settings.

\paragraph{t-SNE visualization.}
For qualitative visualization only, we obtain two-dimensional embeddings using scikit-learn t-SNE~\citep{van2008visualizing}. Pairwise cosine distances are computed from the full normalized feature vectors. The first two principal components provide the initial two-dimensional coordinates, which t-SNE then optimizes. We use perplexity 30, an automatic learning rate, and 1,000 iterations, with a fixed random seed.

\paragraph{Clustering and metrics.}
Let $f_i$ denote the original salient vector of image $i$ and $u_i=f_i/\lVert f_i\rVert_2$ its unit-norm representation. We apply Euclidean $k$-means directly to $\{u_i\}$, using the cluster counts specified above and 10 initializations. Adjusted Rand Index (ARI)~\citep{hubert1985comparing} and Normalized Mutual Information (NMI)~\citep{strehl2002cluster} measure agreement with evaluation labels and are averaged over $k$-means seeds $\{0,1,2\}$, not independent training runs. Tables~\ref{tab:benchmark_main}, \ref{tab:loss_ablation_main}, \ref{tab:oct_stage1}, and~\ref{tab:loss_ablation_full} use this protocol; in Table~\ref{tab:ari_spaces}, Raw denotes it, and PCA-2D and PCA-50 apply PCA to $\{u_i\}$ before $k$-means.

\section{Additional Stage 1 Details}
\label{app:stage1_details}
This section describes the Stage~1 network architectures and training settings, and provides the complete definitions of the objectives summarized in Section~\ref{sec:stage1_method}.

\subsection{Network Architectures}
\label{app:stage1_architectures}
Figure~\ref{fig:cs_encoder_arch} illustrates the architecture used for the common and salient encoders, which map frozen RAE features to the two additive representations. Figure~\ref{fig:discriminator_arch} shows the discriminator, whose shared frozen DINOv1 ViT-S/8 backbone~\citep{caron2021emerging} supplies features to separate background and target expert heads for the swap adversarial objective. This backbone is independent of the DINOv3 encoder being factorized.

\begin{herefigure}
\centering
\includegraphics[width=0.7\linewidth]{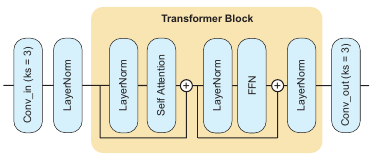}
\caption{\textbf{Common and salient encoder architecture.} Input and output convolutions use kernel size 3. The Transformer block applies layer normalization, self-attention, and a feed-forward network (FFN), with residual connections around the attention and FFN sublayers. The two encoders produce the common and salient representations, respectively. In the salient encoder, both the weights and biases of \texttt{Conv\_out} are initialized to zero.}
\label{fig:cs_encoder_arch}
\end{herefigure}

\begin{herefigure}
\centering
\includegraphics[width=0.7\linewidth]{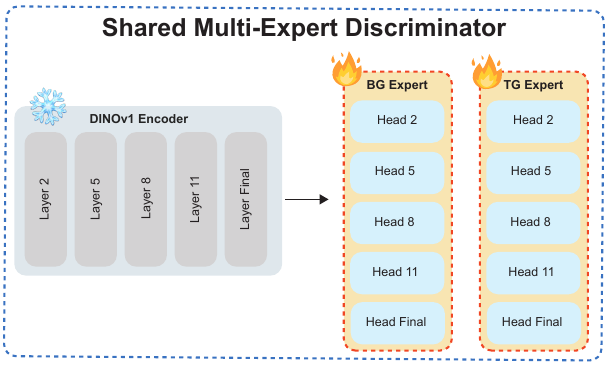}
\caption{\textbf{Shared multi-expert discriminator architecture.} A frozen DINOv1 ViT-S/8 backbone provides features from layers 2, 5, 8, 11, and the final layer to separate background (BG) and target (TG) experts. Each expert contains trainable heads for the corresponding feature levels, enabling group-specific real/fake discrimination. Snowflake and flame symbols indicate frozen and trainable components, respectively.}
\label{fig:discriminator_arch}
\end{herefigure}

\subsection{Stage 1 Training Setup}
\label{app:stage1_training}
\paragraph{Shared optimization settings.}
For both FFHQ and Digits-ImageNet, we use four NVIDIA H200 GPUs with 32 images per GPU, giving a global batch size of 128 (approximately 64 background and 64 target images). The common/salient encoders and trainable discriminator heads use AdamW with the same fixed learning rate of $3\times10^{-5}$, $\beta_1 = 0.9,\beta_2 = 0.999$, weight decay $0.01$, and $\varepsilon = 10^{-6}$. Training uses bfloat16 mixed precision and gradient-norm clipping at 1.0. The DINOv3 backbone and RAE decoder remain frozen, as does the discriminator's feature-extraction backbone; only the common/salient encoders and discriminator heads are optimized.

\paragraph{Dataset-specific settings.}
We train for 200 epochs on FFHQ and three epochs on Digits-ImageNet and use the last saved checkpoint. For FFHQ, we set the target salient $L_1$ weight to $\lambda=0.01$; for Digits-ImageNet, $\lambda=1.0$. The background salient $L_2$ weight is 1.0 for both datasets. For OCT-Kermany, we use two NVIDIA H100 GPUs with 32 images per GPU (global batch size 64) and a class-balanced sampler with 32 normal and 32 diseased B-scans per batch. We otherwise use the shared optimization and frozen-component settings above, train for 60 epochs, and use the last saved checkpoint. The target salient $L_1$ weight is $\lambda=0.01$ and the background salient $L_2$ weight is 1.0.

\subsection{Swap Adversarial Objective}
\label{app:swap_objective}
The generator, consisting of the common and salient encoders, minimizes the non-saturating adversarial objective~\eqref{eq:swap-generator-loss}. The discriminator is trained with the hinge objective
\begin{equation}
\begin{aligned}
\mathcal{L}^{D}_{\mathrm{swap}}
={}&\mathbb{E}_{\hat{x}^b,\hat{x}^t}\!\left[
\ell_{+}\!\left(\Delta_b(\hat{x}^b)\right)
+\ell_{-}\!\left(\Delta_b(\hat{x}^{b}_{\mathrm{fake}})\right)
\right]\\
&+\mathbb{E}_{\hat{x}^b,\hat{x}^t}\!\left[
\ell_{+}\!\left(\Delta_t(\hat{x}^t)\right)
+\ell_{-}\!\left(\Delta_t(\hat{x}^{t}_{\mathrm{fake}})\right)
\right],
\end{aligned}
\end{equation}
where
\begin{equation}
\ell_{+}(s)=\max(0,1-s),
\qquad
\ell_{-}(s)=\max(0,1+s).
\end{equation}
Here, $s$ is the scalar real/fake score predicted by the corresponding discriminator head. The discriminator and common and salient encoders are updated in alternation at each training step.

\subsection{Consistency Objectives}
\label{app:consistency_objectives}
We write $\mathrm{sg}[\cdot]$ for the stop-gradient operator.
\paragraph{Image-space cycle consistency.}
To enforce consistency in the learned representation space for fake background and fake target images, we re-encode the fake images through the frozen backbone and the same common and salient encoders:
\begin{equation}
\hat{z}^{b}=E\!\left(\hat{x}^{b}_{\mathrm{fake}}\right),
\qquad
\hat{z}^{t}=E\!\left(\hat{x}^{t}_{\mathrm{fake}}\right).
\end{equation}
We then decompose the re-encoded representations as
\begin{equation}
\hat{z}^{b}_{c}=E_c(\hat{z}^{b}),\quad
\hat{z}^{b}_{s}=E_s(\hat{z}^{b}),\quad
\hat{z}^{t}_{c}=E_c(\hat{z}^{t}),\quad
\hat{z}^{t}_{s}=E_s(\hat{z}^{t}).
\end{equation}
The cycle objective~\citep{zhu2017unpaired} encourages these re-encoded representations to recover the factors that produced the fake images:
\begin{equation}
\begin{aligned}
\mathcal{L}_{\mathrm{cyc}}
=\mathbb{E}_{x^b,x^t}\Big[
&\left\lVert \hat{z}^{t}_{c}-\mathrm{sg}[z_c^{b}]\right\rVert_2^2
+\left\lVert \hat{z}^{t}_{s}-\mathrm{sg}[z_s^{t}]\right\rVert_2^2\\
&+\left\lVert \hat{z}^{b}_{c}-\mathrm{sg}[z_c^{t}]\right\rVert_2^2
+\left\lVert \hat{z}^{b}_{s}\right\rVert_2^2
\Big].
\end{aligned}
\end{equation}
The first two terms encourage recovery of the donor background  common representation $z_c^{b}$ and the target salient representation $z_s^{t}$ from the fake target image $\hat{x}^{t}_{\mathrm{fake}}$. The last two terms encourage recovery of the target common representation $z_c^{t}$ from the fake background image $\hat{x}^{b}_{\mathrm{fake}}$ while penalizing its salient activations.

\paragraph{Latent-space swap consistency.}
To encourage additive composability and recovery of the constituent representations directly in the normalized representation space, we sample a random permutation $\pi$ for a minibatch $\{(z_c^i,z_s^i)\}_{i=1}^{B}$ and construct
\begin{equation}
z_{\mathrm{cyc}}^{i}=z_c^{i}+z_s^{\pi(i)},\qquad
z_{c,\mathrm{cyc}}^{i}=E_c(z_{\mathrm{cyc}}^{i}),\qquad
z_{s,\mathrm{cyc}}^{i}=E_s(z_{\mathrm{cyc}}^{i}).
\end{equation}
Unlike image-space cycle consistency, this operation does not decode the composed representation or re-run the frozen backbone. The loss encourages the encoders to recover the two factors that formed each composition:
\begin{equation}
\mathcal{L}_{\mathrm{lsc}}
=\frac{1}{B}\sum_{i=1}^{B}\left[
\left\lVert z_{c,\mathrm{cyc}}^{i}-\mathrm{sg}[z_c^{i}]\right\rVert_2^2
+\left\lVert z_{s,\mathrm{cyc}}^{i}-\mathrm{sg}[z_s^{\pi(i)}]\right\rVert_2^2
\right].
\end{equation}
The stop-gradient targets prevent the source representations from moving to match the re-separated outputs. This objective encourages the recovered common and salient representations to be consistent across a wider variety of donors. 

\subsection{Cycle NCE Loss}
\label{app:cycle_nce}
For each paired background and target sample, let $z_c^{b,i}$ and $z_s^{t,i}$ denote the original common and salient representations, and let $\hat{z}_c^{t,i}=E_c(\hat{z}^{t,i})$ and $\hat{z}_s^{t,i}=E_s(\hat{z}^{t,i})$ denote the factors recovered from the re-encoded fake target $\hat{z}^{t,i}=E(\hat{x}^{t,i}_{\mathrm{fake}})$, as in Appendix~\ref{app:consistency_objectives}. Both are therefore recovered from the same swapped image. We globally average-pool and $\ell_2$-normalize all representations:
\begin{equation}
\bar{z}=\frac{\mathrm{GAP}(z)}{\left\lVert\mathrm{GAP}(z)\right\rVert_2}.
\end{equation}
For each salient anchor $\bar{\hat{z}}_s^{t,i}$, the corresponding original salient representation $\overline{z_s^{t,i}}$ is the only positive. The original common representations $\{\overline{z_c^{b,j}}\}_{j=1}^{B}$ and their re-encoded counterparts $\{\bar{\hat{z}}_c^{t,j}\}_{j=1}^{B}$ are negatives. Other salient representations are excluded from the negative set. The common branch is defined symmetrically.

\begin{figure}[tb]
\centering
\includegraphics[width=0.45\linewidth]{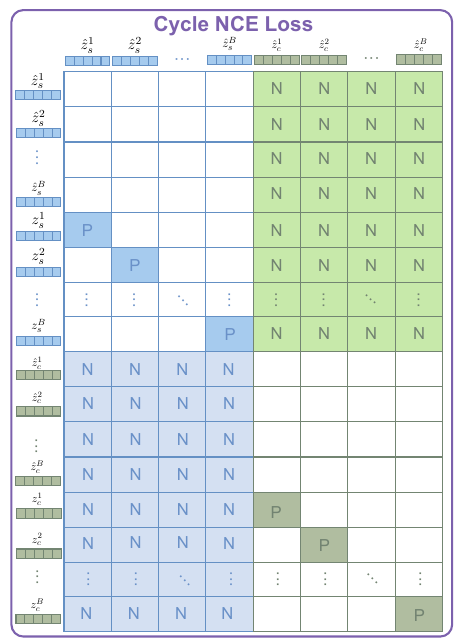}
\caption{\textbf{Cycle NCE pair construction.} ``P'' denotes corresponding same-factor positive pairs, ``N'' denotes cross-factor negative pairs, and blank cells denote excluded same-factor off-diagonal pairs.}
\label{fig:cycle_nce_pairs}
\end{figure}

Figure~\ref{fig:cycle_nce_pairs} visualizes this pair construction. Each re-encoded representation is aligned with its corresponding original representation from the same factor, while original and re-encoded representations from the other factor serve as negatives. Same-factor off-diagonal pairs are excluded.

With temperature $\kappa$, the salient-side loss is
\begin{equation}
\ell_s
=-\frac{1}{B}\sum_{i=1}^{B}
\log
\frac{
\exp\!\left(\bar{\hat{z}}_s^{t,i}\cdot\mathrm{sg}\!\left[\overline{z_s^{t,i}}\right]/\kappa\right)
}{
\exp\!\left(\bar{\hat{z}}_s^{t,i}\cdot\mathrm{sg}\!\left[\overline{z_s^{t,i}}\right]/\kappa\right)
+\displaystyle\sum_{j=1}^{B}\left[
\exp\!\left(\bar{\hat{z}}_s^{t,i}\cdot\mathrm{sg}\!\left[\overline{z_c^{b,j}}\right]/\kappa\right)
+\exp\!\left(\bar{\hat{z}}_s^{t,i}\cdot\bar{\hat{z}}_c^{t,j}/\kappa\right)
\right]
}.
\end{equation}
The common-side loss is
\begin{equation}
\ell_c
=-\frac{1}{B}\sum_{i=1}^{B}
\log
\frac{
\exp\!\left(\bar{\hat{z}}_c^{t,i}\cdot\mathrm{sg}\!\left[\overline{z_c^{b,i}}\right]/\kappa\right)
}{
\exp\!\left(\bar{\hat{z}}_c^{t,i}\cdot\mathrm{sg}\!\left[\overline{z_c^{b,i}}\right]/\kappa\right)
+\displaystyle\sum_{j=1}^{B}\left[
\exp\!\left(\bar{\hat{z}}_c^{t,i}\cdot\mathrm{sg}\!\left[\overline{z_s^{t,j}}\right]/\kappa\right)
+\exp\!\left(\bar{\hat{z}}_c^{t,i}\cdot\bar{\hat{z}}_s^{t,j}/\kappa\right)
\right]
}.
\end{equation}
The final objective is $\mathcal{L}_{\mathrm{cNCE}}=(\ell_s+\ell_c)/2$. The original representations are stop-gradient targets, whereas the re-encoded representations carry gradients. Negatives are gathered across all data-parallel replicas, with gradients retained only for locally computed representations.

\section{Additional Stage 2 Details}
\label{app:stage2_details}
This section expands the Stage~2 formulation summarized in Section~\ref{sec:stage2_method}.

\paragraph{Flow-matching details.}
The salient condition token is concatenated with the noisy latent and time embeddings. Before interpolation, the frozen RAE tokens $z^t\in\mathbb{R}^{N\times C}$ are reshaped into a spatial feature map of shape $C\times H\times W$, where $N=HW$; we keep the symbol $z^t$ for this map. The flow-matching objective is given in Section~\ref{sec:stage2_method}. The time $\tau$ is sampled using the shifted logit-normal schedule of RAEv2~\citep{singh2026improved}, and $\tau_{\min}$ corresponds to \texttt{t\_eps} in Appendix~\ref{app:ffhq_stage2_training}.

\paragraph{Internal guidance and sampling.}
Following the dual-head design of RAEv2~\citep{singh2026improved}, an auxiliary prediction head is attached to an intermediate transformer block and optimized with the same flow-matching objective, giving a second velocity estimate $\hat{v}_\theta^{\mathrm{base}}$. At inference, the full and auxiliary predictions are combined as
\begin{equation}
v^{\mathrm{IG}}
=\hat{v}_\theta^{\mathrm{base}}
+\omega\left(\hat{v}_\theta-\hat{v}_\theta^{\mathrm{base}}\right)
\end{equation}
within the prescribed guidance interval, while $\hat{v}_\theta$ is used elsewhere. This provides internal guidance without requiring a separate unconditional forward pass. On Digits-ImageNet, we additionally apply classifier-free guidance~\citep{ho2022classifier} with the null-token condition (Appendix~\ref{app:ffhq_stage2_training}).

Starting from Gaussian noise, we integrate the flow from $\tau=1$ to $\tau=0$ and reshape the resulting spatial feature map $z_{\mathrm{gen}}$ back into tokens for the frozen RAE decoder:
\begin{equation}
\hat{x}=D\!\left(\operatorname{reshape}(z_{\mathrm{gen}})\right).
\end{equation}
Conditioning on $c_s$ transfers the salient factor of a reference target image, while different noise samples vary the remaining visual content. Setting the condition to the null token instead yields unconditional samples from the target distribution.

\subsection{Stage 2 Training and Sampling Setup}
\label{app:ffhq_stage2_training}

\paragraph{Model and conditioning.}
Both datasets use a Decoupled Diffusion Transformer initialized from RAEv2's EMA-averaged ImageNet DINOv3-L/16 K7 checkpoint, with all parameters fine-tuned. The model predicts the clean latent, the full normalized DINOv3 latent of shape $1024\times16\times16$ given by the K7 multi-layer sum over blocks $\{11,13,\ldots,23\}$. The frozen Stage~1 models are those described in Appendix~\ref{app:stage1_training}. Global max pooling of SAGE's salient latent produces a 1,024-dimensional vector, projected into one condition token with dropout probability 0.1. Both use logit-normal time sampling with parameters $(0,1)$ and \texttt{t\_eps}$=0.05$.

\paragraph{Shared optimization settings.}
Stage~2 trains only on target images, with an effective target batch size of 32 for both datasets. We use AdamW with betas $(0.9,0.95)$, optimizer epsilon $10^{-8}$, zero weight decay, bfloat16 mixed precision, gradient-norm clipping at 1.0, and EMA decay 0.9995. The learning rate warms up linearly to $10^{-4}$ over 0.5 epoch, follows cosine decay to $2\times10^{-5}$ over the first 80\% of training, and remains constant thereafter.

\paragraph{Dataset-specific training settings.}
On FFHQ, we use four NVIDIA H200 GPUs with eight target images per GPU and no gradient accumulation, for 40 epochs (12,480 optimizer steps), with a 156-step warmup and cosine decay ending at epoch 32 (step 9,984). On Digits-ImageNet, we use one NVIDIA H200 GPU with a target micro-batch of 16 and gradient accumulation over two micro-batches, for 20 epochs (400,360 optimizer steps), with a 10,009-step warmup and cosine decay ending at epoch 16 (step 320,288); the remaining four epochs use the final learning rate.

\paragraph{Sampling and evaluation.}
We evaluate the final Stage~2 checkpoints. Both use 50-step Euler sampling and internal guidance scale $\omega=1.78$ active over $\tau\in[0.10,1.0]$. Digits-ImageNet additionally uses classifier-free guidance scale 1.5, as does its Raw DINOv3 GMP baseline in Table~\ref{tab:stage2_generation}; FFHQ uses none.

\subsection{Within-Condition Diversity Evaluation}
\label{app:within_condition_diversity}
We investigate how much a Stage-2 model varies its output under a fixed conditioning image as the initial noise changes by computing the Vendi Score~\citep{friedman2023vendi} from the eigenvalues of the CLS output of DINOv2-L~\citep{oquab2023dinov2} (\texttt{facebook/dinov2-large}). Diversity must be measured within a reference for non-target attribute variation to be visible: scored across a set spanning many references, the variation among the references themselves enters the number, and a model that copies each reference would appear diverse on inherited variation alone. We report these results in Table~\ref{tab:stage2_generation}.

\paragraph{Reference images and noise.}
We draw reference images by stratified random sampling from the target validation images. For FFHQ, we sample 30 images from each manually verified eyewear subtype---reading glasses and sunglasses---for 60 images total; for Digits-ImageNet, we sample 10 images for each digit class (100 images total). Each reference image is paired with $K=16$ independently generated noise samples. SAGE and Raw DINOv3 GMP use the identical reference lists and initial noises, pairing generated samples image by image. SAGE pools the Stage~1 salient representations with global max pooling (GMP), whereas Raw DINOv3 GMP bypasses the salient encoder and pools the original Stage~1 DINOv3 features with GMP. The two Stage~2 models are trained separately with the same training budget.

\paragraph{Sampling configuration.}
We use each Stage~2 model's own EMA weights (decay 0.9995), 50-step Euler sampling, internal guidance scale $\omega=1.78$ active over $\tau\in[0.10,1.0]$, bfloat16 inference, and $256\times256$ output resolution for both conditions. FFHQ uses classifier-free guidance scale 1.0 and Digits-ImageNet uses 1.5. These settings match the Stage~2 main-table evaluation.

\paragraph{Feature extraction and metrics.}
We extract the LayerNorm pooler (CLS) output of DINOv2-L. Each image is resized to $224\times224$, and its feature is $\ell_2$-normalized. For the $K=16$ generated samples of one reference image, let $f_i$ denote the normalized feature of sample $i$ and form the cosine-kernel matrix $K_{ij}=f_i^\top f_j$. We compute the Vendi Score~\citep{friedman2023vendi} from the eigenvalues $\{\lambda_i\}$ of $K/16$:
\begin{equation}
\mathrm{VS}=\exp\left(-\sum_i \lambda_i\log \lambda_i\right).
\end{equation}
The score ranges from 1, for 16 identical samples, to 16, for pairwise orthogonal features, and can be interpreted as the effective number of perceptually distinct images. We also compute the cosine similarity between every generated sample and its own reference-image feature, averaged over the 16 samples, to quantify reference-copying tendency.  

\paragraph{Real-data diversity reference.}
For every reference image, we randomly draw 16 target validation images from the same subtype and compute the Vendi score with the same feature extractor. This reference estimates the within-subtype diversity available in the real validation data; it is reported as a data-diversity reference rather than as an absolute bound.

\section{Limitations}
\label{app:limitations}
\paragraph{Discovery versus complete disentanglement.}
Background/target labels do not uniquely determine how correlated attributes are split between the two factors. The common representation retains linearly decodable digit information (probe accuracy $0.336$, comparable to $0.328$ for the unfactorized latent; Table~\ref{tab:loss_ablation_full}), although common-only decoding removes the digit (Fig.~\ref{fig:factor-decoding-benchmark}). SAGE therefore yields a clean salient factor rather than an exclusive assignment of target information to it.


\paragraph{Limitations on biomedical images.}
SAGE inherits the representation and reconstruction priors of its frozen encoder and decoder, which are pretrained on natural images. Reconstruction on OCT-Kermany is accordingly less faithful than on natural images (SSIM $0.448$ versus $0.571$ on Digits-ImageNet and $0.723$ on FFHQ; Table~\ref{tab:oct_stage1}). Because SAGE keeps the autoencoder frozen, a representation autoencoder pretrained on biomedical images could replace RAEv2 without changing the method.

\end{document}